\documentclass{article}

     \PassOptionsToPackage{numbers, compress}{natbib}

     \usepackage[preprint]{neurips_2020}

\usepackage[utf8]{inputenc} 
\usepackage[T1]{fontenc}    
\usepackage{hyperref}       
\usepackage{url}            
\usepackage{booktabs}       
\usepackage{amsfonts}       
\usepackage{nicefrac}       
\usepackage{microtype}      

\usepackage{amsmath, amssymb, amsfonts, amsthm}
\usepackage[english]{babel}
\usepackage{graphicx}
\usepackage{fancyhdr}
\usepackage{color}
\usepackage{hyperref}
\usepackage{bbm}
\usepackage{graphbox}
\usepackage[capitalize]{cleveref}
\usepackage{array}
\usepackage{tabularx}

\usepackage{xcolor}
\usepackage{listings}
\newcolumntype{L}[1]{>{\raggedright\let\newline\\\arraybackslash\hspace{0pt}}m{#1}}
\newcolumntype{C}[1]{>{\centering\let\newline\\\arraybackslash\hspace{0pt}}m{#1}}
\newcolumntype{R}[1]{>{\raggedleft\let\newline\\\arraybackslash\hspace{0pt}}m{#1}}

\usepackage{bbold}			

\renewcommand*\arraystretch{1.3}

\graphicspath{{./plots_compressed/}}

\newcommand{\pd}{\textit{PredDiff}}
\newcommand{\ig}{\textit{Integrated Gradients}}
\newcommand{\iga}{\textit{IG}}

\newcommand{\imagenet}{ImageNet}
\newcommand{\celeba}{CelebA}

\newcommand{\norm}[1]{\left\lVert#1\right\rVert}
\newcommand{\R}{\mathbb{R}}

\title{Sparse Subspace Clustering for Concept Discovery (SSCCD)}

\author{
  Johanna Vielhaben\\
  Explainable Artificial Intelligence Group\\
  Fraunhofer Heinrich-Hertz-Institute\\
  10587 Berlin, Germany \\
  \texttt{johanna.vielhaben@hhi.fraunhofer.de} \\
   \And
   Stefan Bl\"ucher \\
   Machine Learning Group \\
   TU Berlin \\
   10587 Berlin, Germany \\
   \texttt{bluecher@tu-berlin.de} \\
   \AND
  Nils Strodthoff \thanks{Work partially done at Fraunhofer Heinrich-Hertz-Institute.} \\
  Division AI4Health \\
  Oldenburg University  \\
  26129 Oldenburg \\
  \texttt{nils.strodthoff@uol.de}
}

\begin{document}

\maketitle

\begin{abstract}
Concepts are key building blocks of higher level human understanding.
Explainable AI (XAI) methods have shown tremendous progress in recent years, however, local attribution methods do not allow to identify coherent model behavior across samples and therefore miss this essential component.
In this work, we study concept-based explanations and put forward a new definition of concepts as low-dimensional subspaces of hidden feature layers.
We novelly apply sparse subspace clustering to discover these concept subspaces. 
Moving forward, we derive insights from concept subspaces in terms of localized input (concept) maps, show how to quantify concept relevances and lastly, evaluate similarities and transferability between concepts.
We empirically demonstrate the soundness of the proposed Sparse Subspace Clustering for Concept Discovery (SSCCD) method for a variety of different image classification tasks.
This approach allows for deeper insights into the actual model behavior that would  remain hidden from conventional input-level heatmaps.
\end{abstract}

\section{Introduction}
\label{sec:intro}
In neuroscience, concepts are defined as flexible, distributed representations comprised of modality-specific conceptual features and are important building blocks of human cognition \cite{cortex2012}. Like neuroscience sheds light on human mental processes, the subfield of Explainable AI in machine learning develops methods that provide insights into the decision processes in particular inside deep neural networks, see \cite{covert2021explaining,lundberg2017unified,Montavon2018,Samek2021} for reviews.

Here, we are concerned with \textit{local} interpretability methods that provide attributions for individual samples, which is desirable for most applications. However, to acquire a higher-level understanding of the model, the user needs to aggregate numerous samples into a common explanation. Not only is the human ability for this limited, but the summarization is also prone to human confirmation bias \cite{tcav2018}. Even leaving these issues aside, the resulting insights do not reflect the actual reasoning structure hidden for example in the feature layers of a model.
This urges for novel \emph{local} and \emph{concept-based} interpretability methods, that explore shared structures in the decision process of the neural network across several examples. We stress already at this point that our primary motivation is to reveal coherent, discriminative structures exploited by the model rather than achieving best accordance with concepts identified by humans.

\begin{figure}[h!]
	\centering
	\def\arraystretch{0.5} 		
	\setlength\tabcolsep{1pt}		
	\small
	\begin{tabular}{cccc}
		 \includegraphics[width=0.12\linewidth]{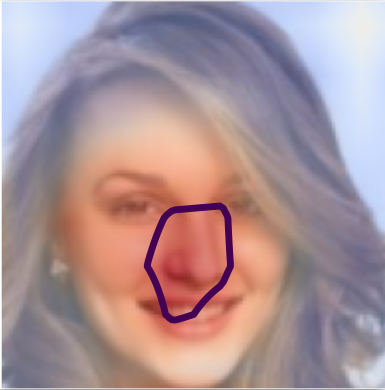}	&
 		\includegraphics[width=0.12\linewidth]{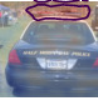} 		&
		\includegraphics[width=0.12\linewidth]{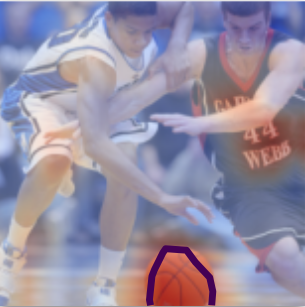}		&
		\includegraphics[width=0.12\linewidth]{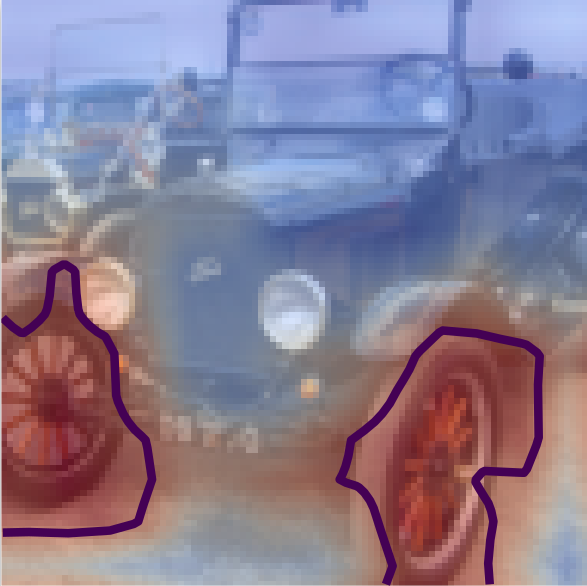}	\\
		ResNet50&	ResNet50	& 	VGG16	& VGG16		\\
		 \celeba	&		\imagenet & \imagenet & \imagenet
	\end{tabular}
	\caption{SSCCD successfully discovers concepts for a variety of datasets and models.
	}
	\label{fig:highlights}
\end{figure}

The common aspect of exisiting concept-based interpretability methods, such as \cite{ace2019,tcav2018,yeh2020}, is the fact that they understand concepts as single vectors in the space of intermediate hidden feature layers and are in this sense meaningful to the model.
It lies in the nature of a concept as an abstract representation that its expressions can be diverse. Therefore, it is surprising that the definition of concepts as one-dimensional feature subspaces has so far not been challenged. We novelly argue for concepts as low-dimensional subspaces. Multiple feature dimensions enable to fully capture the \textit{intra-concept diversity} described above (\textit{benefit 1}).
To discover these low-dimensional concept subspaces in feature space we build on the successful application of sparse subspace clustering in input space \cite{elhamifar2013sparse}.
The approach stands out from many of its competitors as it does not require reinforcing the interpretability of concepts (e.g. through prior superpixel segmentation, regularizer to enforce the dissimilarity of concepts etc.\ ) and thus reflects the inner workings of the model as faithfully as possible (\textit{benefit 2}). 
Additionally, we characterize the identified concept subspaces by a basis and propose similarity measures between them. In contrast to generic, untargeted concepts for the whole classifier, our approach allows extracting class-specific concepts, as well as concepts based on specific combinations of classes. Altogether, this enables a quantitative comparison of concepts, the transferability to different classes, and thus a comprehensive model understanding. 

Every concept interpretability method comes with a number of sub-components: (a) concept discovery (b) mapping from feature space to input space (c) relevance quantification. We stress at this point that any concept interpretability method has to decide on these sub components either explicitly or implicitly. 
As mentioned above, we see an appropriately modified sparse subspace clustering as an ideal fit for the identification of concept subspaces (a).
For (b), we follow \cite{selvaraju2020grad} and translate feature level concept maps to the input level by simple bilinear interpolation.
In our view, alternative mappings, which are usually based on receptive fields, often overestimate the area of impact.
Finally, concerning (c), we put forward a framework that allows inferring the concept relevance with any interpretability method either anchored in input or feature space. This leads to a robust method that works across different datasets and model architectures, see \Cref{fig:highlights}.
To summarize, our main contributions are the following:
\begin{itemize}
	\item Novel concept definition as low-dimensional feature subspaces
	\item Sparse subspace clustering for concept discovery
	\item Relevance quantification for concepts via user-defined attribution method 
	\item Insights from a quantitative similarity measure between concept subspaces 
\end{itemize}

\section{Sparse Subspace Clustering for Concept Discovery (SSCCD)}
We start by motivating our definition of concepts which form the building blocks of a deep neural network's reasoning process. For definiteness, we focus on convolutional neural networks (CNN) in this section and comment on extensions to other architectures below. Our approach is based on the following principles, which align with the aspects put forward in the introduction:

(1) Concepts should be \textit{meaningful to the model}, i.e.\ should be inherently \textit{tied to hidden feature layers} to ensure that the concepts correspond to intermediate processing steps within the network. (2) Concepts should allow to capture the intra-concept diversity that is inherent to the typically abstract nature of concepts. 
In this work, we argue that the until now uncontested definition \cite{ace2019,tcav2018,yeh2020} of concepts as single directions, i.e.\ one-dimensional feature subspaces is generally not enough and postulate to define \textit{concepts as linear subspaces} in feature space. For practical considerations, we require these subspaces to be \textit{robust against noise and outliers}.

In addition, we believe that there are further desirable properties that a concept discovery framework (CDF) should satisfy:

(3) A CDF should allow to \textit{spatially localize} the concept in feature or input space, see the discussion on \textit{concept maps} below, whereas the definition of the concept itself should not explicitly make reference to the spatial location within the feature layer.  
(4) A CDF should provide a \textit{relevance quantification} of a concept in terms of its impact on the classification decision.
(5) A CDF should allow to assess the \textit{similarity} between different concepts, both within and across different classes. Given a sensible concept definition, concepts should be \textit{transferable} between samples from semantically similar classes.

\begin{figure}
	\centering
	\includegraphics[width=\linewidth]{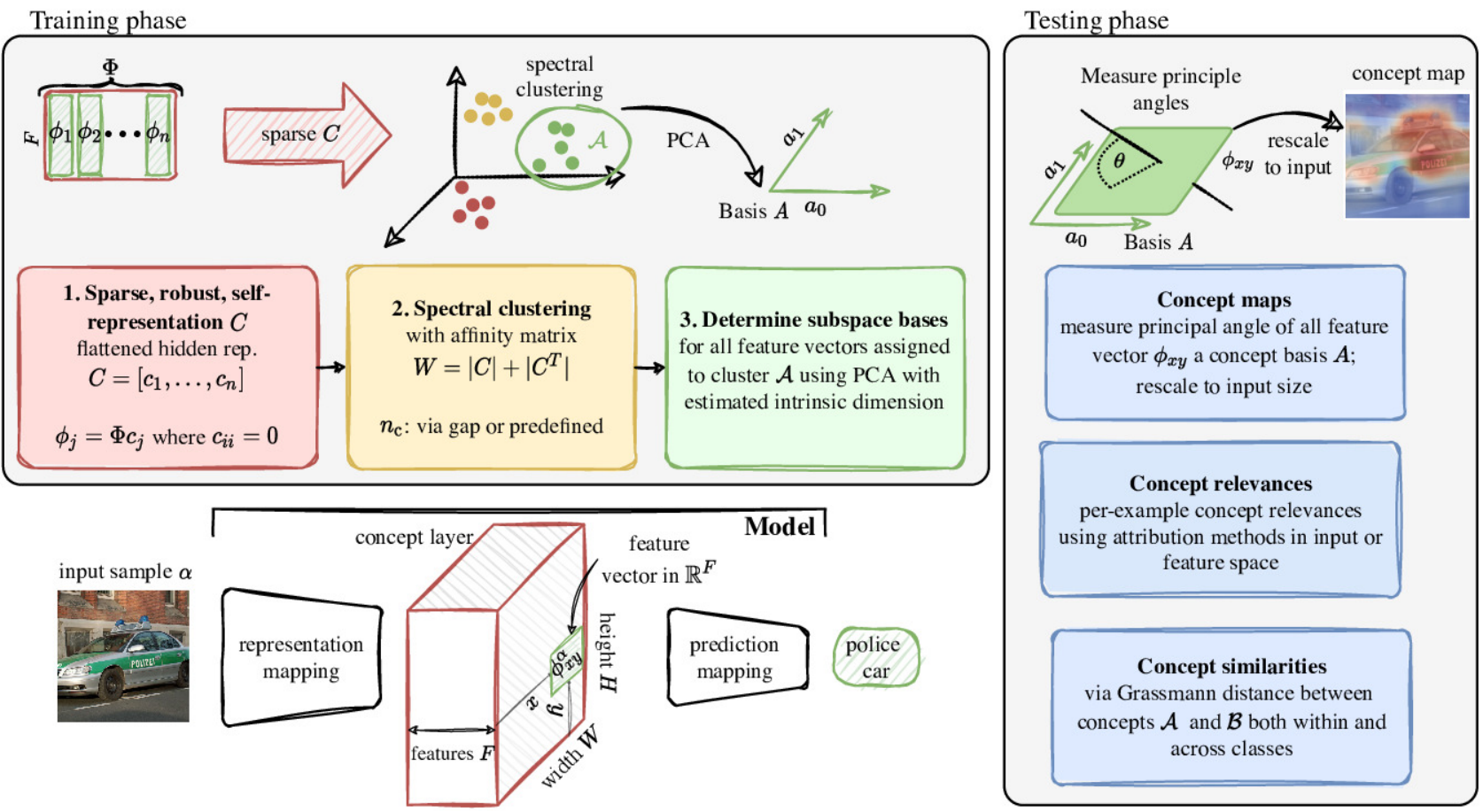}
	\caption{Schematic illustration of the SSCCD-approach for concept discovery.
	}
	\label{fig:schematic}
\end{figure}

\paragraph{Concept definition} As a first step, we select a set of samples $S$, for which we aim to discover concepts.
	We pose no restrictions on the samples in $S$, i.e. the user can decide for class-specific samples/concepts or up to all training samples to obtain completely class-unspecific concepts.
Our definition is then based on the hidden representations $h(\alpha \in \R^{H \times W \times F })$ of input samples $\alpha \in S$ with width (of the feature layer) $H$, height $W$ and number of features $F$. We spatially deconstruct the feature maps $h(\alpha)$ and obtain a feature vector $\phi^\alpha_{xy} \in \R^F$ for each location $(x,y) \in \{1,\ldots,H \} \times \{1,\ldots,W\}$.
We define a concept as a robust subspace of the $F$-dimensional feature space.

Note that this definition does not include any constraints on the spatial coherence of concepts nor does it include additional constraints such as penalizing similar clusters as in \cite{yeh2020}. In this way, the concept discovery process does not obfuscate the structures the model is exploiting. Concepts are as \emph{true to the model} as possible (\textit{benefit 2}). To identify concept subspaces we draw on the rich body of literature on \textit{sparse subspace clustering} (SSC) \cite{elhamifar2013sparse,soltanolkotabi2012geometric,you2016scalable,you2016oracle}.  As nicely laid out in \cite{elhamifar2013sparse}, 
SSC is ideally suited to identify clusters of linear subspaces and provides a number of advantages over standard clustering algorithms, which are directly applied to the data: SSC does not take advantage of the spatial proximity of the data e.g.\ to characterize clusters via a centroid, it can be implemented robustly against noise and outliers and does not require to specify the different cluster dimensionalities in advance. In the following, we describe our method, which we split into training and test phase.

\subsection{Training: Concept discovery}
During training, concepts are discovered through the \textit{Sparse Subspace Clustering for Concepts Discovery} (SSCCD) procedure, see also \Cref{fig:schematic} for a schematic representation:

\paragraph{1. Concept-determining self-representation}
We compute sparse self-representations $C$ for a random sub-collection of $n\leq N \cdot  H \cdot W$ feature vectors $\{\phi^\alpha_{xy}\}$ sampled from $S$.
Here, the term self-representation refers to a coefficient matrix that expresses each sample as a linear combination of all other samples. More specifically, using the notation from \cite{elhamifar2013sparse}, given the feature vectors $\Phi=[\phi_1,\ldots,\phi_n]\in \R^{F \times n }$, we identify a sparse coefficient matrix $C=[c_1,\ldots,c_n]\in \R^{n \times n}$ such that
\begin{equation}
	\label{eq:selfrep}
	\phi_j = \Phi c_j\,\,\text{where}\,\,c_{ii}=0.
\end{equation}

The particular kind of sparsity constraints that are imposed on \Cref{eq:selfrep} and how it is optimized depends on the chosen SSC algorithm. Here, we use elastic net subspace clustering \cite{you2016oracle}, which is robust against noise and scales well for large sample sizes. In all our experiments, we fix the hyperparameter $\gamma$, which balances sparsity vs. robustness, to $\gamma = 10$.

We remove outliers based on the $\ell_1$-norm as in \cite{soltanolkotabi2012geometric}, where we empirically fix the percentile threshold to 0.75 and re-fit the sparse self-representation for the remaining elements.
Another scalable alternative to the elastic net clustering is orthogonal matching pursuit (OMP)\cite{you2016scalable}, which is, however, not robust against noise and does not allow for outlier removal via thresholding. 
Finally, the original sparse subspace clustering method from \cite{elhamifar2013sparse} is robust against noise and outliers but does not scale to large datasets. The particular robustness and scalability properties make elastic net subspace (with thresholding) an ideal choice for the first step of our concept discovery method.
\paragraph{2. Spectral clustering} We perform spectral clustering with the affinity matrix $W=|C|+|C^T|$, which encodes the similarity of two feature vectors according to their self-representations. 
We determine the number of clusters $n_c$ either via the largest gap in the spectrum of the Laplacian \cite{vonLuxburg2007tutorial} or use a predetermined value. This step assigns every input feature $\phi_i$ to a particular cluster $\mathcal{C}_1,\ldots,\mathcal{C}_{n_c}$ or to the set of outliers.

\paragraph{3. Identifying subspace basis elements}
Next, we turn to the identified concepts, which are defined via the corresponding cluster members, and construct a low-dimensional basis as we aim to characterize concepts via subspaces rather than cluster members.
Concretely, for a given cluster $\mathcal{A}$, we aim to identify a basis $A$ that robustly covers $\mathcal{A}$. 
To this end, we apply principal component analysis (PCA) and determine the intrinsic dimension of the subspace using a heuristic proposed by \cite{fukunaga1971algorithm} as implemented in \cite{bac2021scikitdimension}. The PCA components up to the intrinsic dimension then serve as a basis for the subspace~$A$.

We stress that the overall methodology is applicable beyond CNNs. In particular, one can decompose feature representations of any model based on SSCCD. 
However, human identification of the identified concept subspaces requires the ability to connect spatial locations in the feature layer with spatial locations in the input space. Due to the local nature of convolutions, this can be achieved by upsampling the spatial dimensions, as described for \textit{concept maps} below.
Clearly localized attention maps \cite{caron2021emerging} show that locality is induced via skip-connections in transformer architectures.

\subsection{Test: Insights from concept subspaces}
To test the discovered concepts, we evaluate how the model employs them to classify (new and unseen) samples. We visualize where the model identified a concept in input space by \textit{concept (expression) maps} and quantify its \textit{relevance} towards the prediction. Additionally, we quantify the relation between concept subspaces based on a measure for \textit{concept similarities}.
The central tool here are principal angles $\theta_i^{AB}$ \cite{Jordan1875} ($i=1,\ldots,\min(\text{dim}\,A,\text{dim}\,B)$) between two linear subspaces, which are defined recursively via
\begin{equation} 
	\cos \theta_i^{AB}= \underset{a\in A, b\in B}{\max} \tfrac{a^T b}{\norm{a}_{\ell_2} \norm{b}_{\ell_2}} =: \tfrac{a_i^T b_i}{\norm{a_i}_{\ell_2} \norm{b_i}_{\ell_2}}\,,
\end{equation}
 where the maximum is taken subject to the orthogonality constraints $a^T a_j=0$ and $b^T b_j=0$ for $j=1,\ldots,i-1$. 

 \paragraph{Concept (expression) maps}
 We can use the principal angle $\theta_1^{\phi^\alpha_{xy} A}$ to characterize the relation between a feature vector $\phi^\alpha_{xy}$ (i.e.\ a one-dimensional subspace) of the $\alpha$-sample and a concept $\mathcal{A}$.
 If we apply this across all spatial dimensions~$(x, y)$, we can define a \textit{concept map} that spatially resolves the similarity of the feature layer $\phi^\alpha_{xy}$  to the concept, i.e.\ $\theta^{\alpha A}_{xy} := \theta^{\phi^\alpha_{xy}A}_1$. Here, high similarity (i.e. highly activated concepts) corresponds to a small principal angle $\theta^{\alpha A}_{xy}$.
 Due to the local nature of convolutional layers, we next leverage bilinear upsampling to resize the concept map to the input image size.
 This enables to represent the meaning of concepts in input space and allows to overlay images with color-coded concept maps.
 To guide the eye, we include contour lines that are defined via the median of all principal angles belonging to a particular concept, i.e.\ we define a concept-specific \textit{threshold angle} via $\delta^A=\text{Median}(\{\theta_1^{\phi A}| \phi \in \mathcal{A}\})$.
 More precisely, the median is taken with respect to all features vectors which belong  to a particular cluster $\mathcal{A}$ during the concept discovery step 1.
 This provides a way of obtaining hard concept assignments in input or feature space without the necessity of introducing an additional hyperparameter.
 Finally, we can define a \textit{sample-concept proximity} measure $\delta^{\alpha A}=\text{min}_{xy} ( \theta^{\alpha A}_{xy} )$ as the minimal principle angle between a concept subspace $A$ and all feature vectors of the $\alpha$-sample.

 \paragraph{Concept relevances}
 
Finally, we are not only interested in the discovery of concepts as specific structures in the model's feature space, but moreover, we also desire to quantify the concept's relevance for the classification decision on a per-example basis.
For SSCCD this can be achieved by an attribution method either based in input or feature space.

To this end, we use the threshold angles defined above to convert concept maps into binary masks to identify regions in input/feature space that are associated with a particular concept.
Then the relevance of these concept masks can be quantified using an established attribution methods.
As an advantage, our novel SSCCD-approach can be combined with any attribution method in input space or alternatively directly in the feature space.
For definiteness, we consider \pd{} \cite{bluecher2021preddiff,Sikonja2008Explaining} and \ig{} \cite{sundararajan2017axiomatic} in this work.
In the case of \ig{}, we sum up the individual pixel relevances, in contrast for \pd{}, we directly obtain superpixels attributions.
Like \cite{li2021} we first evaluate instance-wise relevance scores and later aggregate these to class-wise concept relevance.

 \paragraph{Concept similarities}
 To quantitatively characterize the similarity between two concepts $\mathcal{A}$ and $\mathcal{B}$, we use their Grassmann distance \cite{hamm2008subspace}, which is defined as $d^{AB}=\sqrt{(\theta_1^{AB})^2+\ldots + (\theta_{\text{min}(\text{dim}\,A,\text{dim}\,B)}^{AB})^2}$.
 This allows comparing the similarity of concepts within a given class or across classes regardless of the concept subspaces' dimensionality.

\section{Related Work}

ACE \cite{ace2019} uses a superpixel segmentation algorithm and $k$-means clustering to identify class-specific concept candidates for TCAV \cite{tcav2018}.

The concept discovery scheme of ACE has several shortcomings: 
The segmentation, i.e. the candidate concept patches, is model-independent and thus segments are not necessarily meaningful as perceived by the model.

To enable clustering of intermediate CNN activations, segments are resized and mean padded to the original input shape. This leads to artificial, off-manifold samples with potentially distorted aspect ratios and discards the overall scale information.

Finally, ACE relies on multiple heuristics to discard segments/clusters both before and after $k$-means clustering.
In contrast, spectral clustering in SSCCD is coherently based on hidden model representations without relying on additional pre- or postprocessing.
Similar limitations apply to methods that rely on ACE-discovered labeled concepts, like \cite{li2021}, which uses Shapley values for concept importance, and \cite{wu2020}, which occludes particular neurons for neuron-wise relevances and transforms them into concept importances via concept classification.

ICE \cite{ice2021} defines concepts as directions in feature space. Technically, this is achieved via dimensionality reduction techniques applied to concatenated flattened feature maps.
ICE measures the importance of its class-wise concepts using TCAV.
Other methods learn concept vectors and a mapping to feature space either for all classes simultaneously (ConceptSHAP \cite{yeh2020}) or for each class separate (MACE  \cite{kumar2021}, PACE \cite{kamaskhi2021}).
Importantly, each method defines a custom measure for concept importance which is only applicable within the respective framework.
All these concept importance measures are based on approximations of the original model. %
In contrast, SSCCD can build on any (established) attribution method both in input or feature space.
Our results indicate that concepts are multi-dimensional in feature space.
ConceptShap, MACE and PACE need to account for this via additional regularizers enforcing dissimilarity.
SSCCD in comparison can identify diverse concept subspaces in a fully unsupervised way.

There is a complementary line of work where one tries to analyze CNN hidden representations in combination with special \emph{concept-annotated} datasets.
Network Dissection \cite{bau2017network} investigates the alignment of human-understandable concepts and particular single hidden features (neurons).
\cite{fong2018} extended this by allowing concepts to be represented by combinations of neurons.
Alternatively, \cite{zhou2018} projects model representations onto an interpretable basis, which is learned from a pixel-level segmented and annotated dataset.

\section{Results}
We limit our experiments to natural images and CNNs and defer extensions to other datasets and architectures to future work. We use two popular, large-scale image recognition datasets, \imagenet \cite{deng2009imagenet} and the CelebFaces Attributes Dataset (\celeba{}) \cite{liu2015faceattributes}. As for \imagenet, we use pretrained VGG16 \cite{simonyan2014very} and ResNet50 \cite{he2016deep} model as provided by \textit{torchvision}. On \celeba{}, we train a ResNet50 model from scratch on the full set of 40 attributes using a binary crossentropy loss. The model reaches a macro AUC of 0.92 on the \celeba{} test set.

In the following, we showcase the abilities of SSCCD for the different datasets and model architectures and derive a number of new insights from them. We stress already at this point that all presented results are strictly based on test set samples whereas the concepts itself are discovered on the training dataset.

Our main analysis tool for qualitative assessment of the discovered concepts are concept maps. In the following, the rows of all concept maps, e.g. \Cref{fig:pca baseline}, are sorted according to their average concept relevances. 
Here, we used \ig{} to obtain attributions and calculate the per-sample concept relevances via summing all attributions over the thresholded (red) regions of the concept maps in input space. 
We further assessed the effect of this relevance quantification method. Firstly, we replaced \ig{} with \pd{} and secondly, measure attributions in feature space directly (for both methods), see supplementary material for details. 
It reassuring that, despite differences in the numerical values, the concept ranking based on these importances is largely consistent across all methods. Lastly, we used a predefined number of five concepts for all of the following experiments.

\subsection{Impact of multi-dimensionality} \label{sec:mutlidim}
First, we investigate the advantage of multidimensional concept subspaces as given by SSCCD's concept definition (\textit{benefit 1}) over concept vectors, i.e.\ one-dimensional subspaces, used in the literature. 
If one restricts to one-dimensional concept vectors only, a different concept discovery method is appropriate. Straightforward candidates are dimensionality reduction methods such as PCA or NNMF, which are directly applied in feature space.

\begin{figure}
	\centering
		\def\arraystretch{0.9} 		
	\setlength\tabcolsep{2.5pt}		
		\begin{tabular}{lrlcr} 
			\multicolumn{2}{c}{high intra-class variance} 	&	&			\multicolumn{2}{c}{low intra-class variance} \\  \cline{0-1}   \cline{4-5}  
			SSCCD \rule{0mm}{3.5mm} 	& 		PCA/ICE-baseline   &		&	SSCCD		& 		PCA/ICE-baseline \\ 
			\multicolumn{2}{c}{\includegraphics[width=0.475\columnwidth]{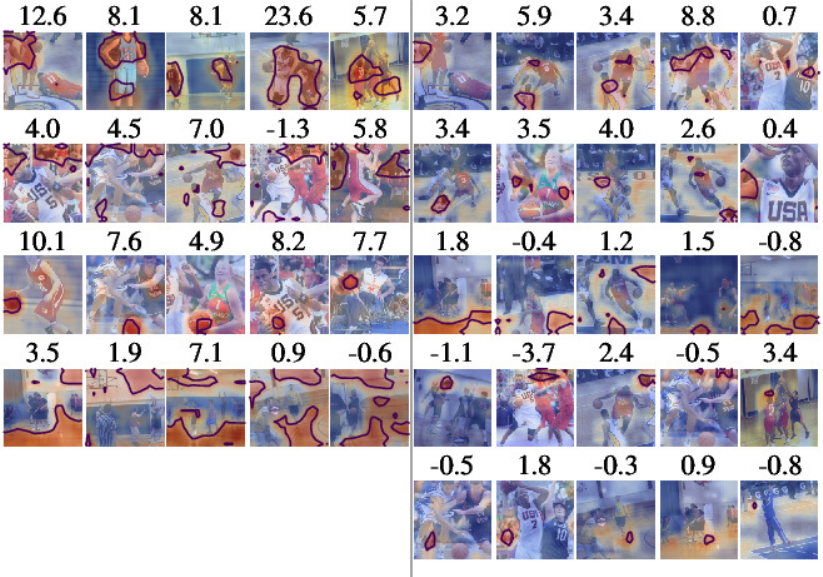}}	& &
			\multicolumn{2}{c}{\includegraphics[width=0.475\columnwidth]{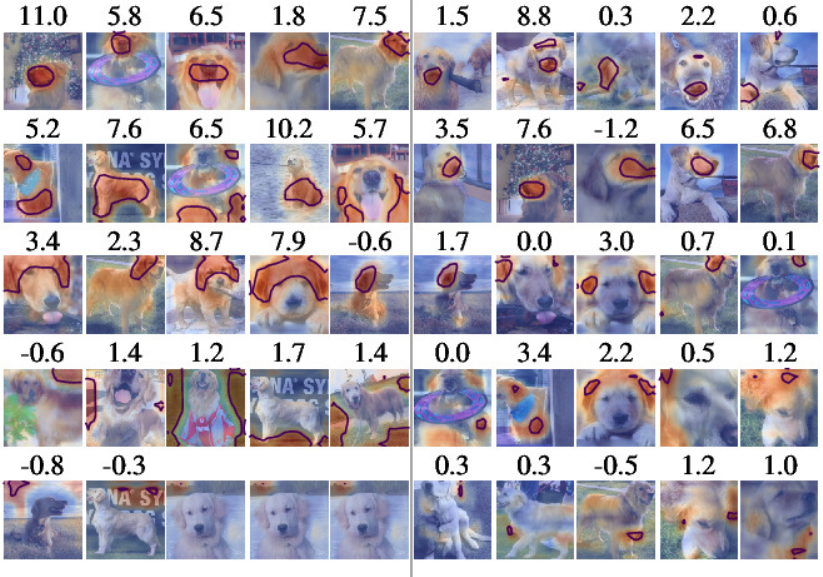}}	\\
		\end{tabular}
	 \caption{Concept maps for basketball class (left) and golden retriever class (right) and a VGG16 model (final convolutional layer). For concept maps, red regions denote high \emph{concept proximity} and lines indicate the concept-dependent thresholds $\delta^A$. We compare multi-dimensional SSCCD concepts
	 to one-dimensional concepts obtained via direct application of PCA in feature space as used by ICE \cite{ice2021}. 
	 	The fifth SSCCD concept of the basketball class was not activated (below threshold) on the whole test set and was therefore omitted. Subtitles indicate the per-sample concept relevances (via \ig{} in input space). Rows are sorted by average concept relevance and the closest proximity samples are shown.}
	 \label{fig:pca baseline}
\end{figure}
In \Cref{fig:pca baseline}, we compare SSCCD concept maps to those obtained from applying PCA directly in feature space, which corresponds to one of the concept discovery methods used by ICE \cite{ice2021}. Different from ICE, we choose a threshold relative to the maximum \textit{sample-concept proximity} per concept across all training samples, using a factor of $1.2$, for the PCA approach. This corresponds to $\delta^A$  for SSCCD.
We focus on the basketball class and the golden retriever class as examples with high and low \textit{intra-class variance}, respectively, see the supplementary material for further class examples. Samples from a class with high \textit{intra-class variance} are more diverse than samples from a class with low \textit{intra-class variance}.
The resulting basketball concepts demonstrate the shortcomings of the one-dimensional PCA approach. 
SSCCD concepts are clearly localized and identifiable (shirt, heads, ball, gym floor/wall).
In contrast, PCA basketball concepts are less localized and often incoherent. In particular, the ball itself is not even recognized as a concept.
For diverse (but coherent) concepts such as colorful shirts or backgrounds, the PCA approach is forced to create separate concept vectors. 
SSCCD, however, can account for this via a common multidimensional concept subspace, which demonstrates \textit{benefit 1}. In accordance with expectations, the PCA approach tends to show most satisfying results for the simpler golden retriever class where samples and hence concept expressions are more uniform than for the basketball class. Still, the PCA approach identifies two separate directions for the nose concept. 

\subsection{Dependence on model architecture}
\begin{figure}
\centering
\hspace{.2cm} VGG16 \hspace{2.6cm} ResNet50 \\
\includegraphics[width=0.5\linewidth]{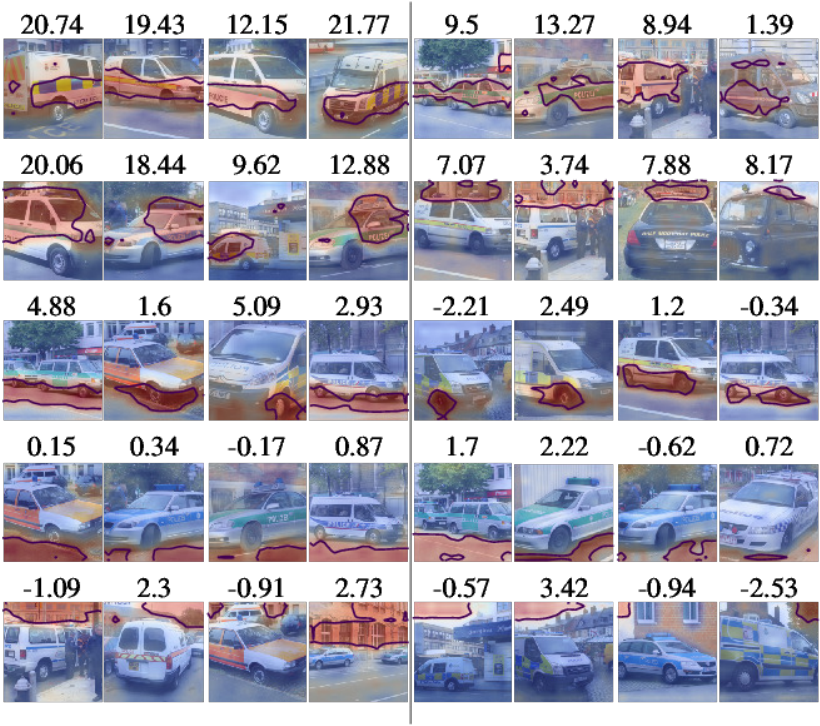}
 \caption{SSCCD concept maps with per-sample concept relevances (via \ig{} in input space) for the police van class from \imagenet. 
 	Each row corresponds to a different concept sorted by average concept relevance.
 	For easier concept identification, we show samples with the smallest sample-concept proximity. 
 	Random selections are presented in the supplementary material.
  	VGG16: last feature layer in the last block. 
	 ResNet50: last feature layer in the second-last block.}
 \label{fig:vgg_vs_imagenet}
\end{figure}
To demonstrate that the approach is applicable irrespective of the particular model architecture, we show concepts for a VGG16 and a ResNet50 model in \Cref{fig:vgg_vs_imagenet}, where we specialize to the police car class. We defer the reader to the supplementary material for alternative \imagenet{} classes.
For comparability, both feature layers in \Cref{fig:vgg_vs_imagenet} were chosen such that they have the equal spatial size $14\times 14$.
Alternative feature layer choices are presented in the supplementary material.

For VGG16,\,SSCCD identifies livery~(8.0), windows~(7.8), tires/underbody~(3.7), street~(0.5), buildings~(0.3) as most important concepts. Here, values in brackets denote the average concept relevance as measured on all test samples. 
For a ResNet50 model we obtain livery/windows~(5.4), emergency lights~(3.1), tires/ underbody~(1.2), street~(0.3), sky/background~(0.02) as most relevant concepts.
Interestingly, both models identify similar concepts and also their ranking, based on concept relevances, agrees.
For ResNet50, the secondly ranked concept of emergency lights stands out, as it was not identified among the VGG concepts.
This provides a hint towards different strategies followed by the two networks. Importantly, such insights are hard to obtain via raw attribution-based explanation strategies. 
Note further, that in both cases also typical background structures were identified as coherent concepts. 
However, these concepts are not informative for the classification task and consequently assigned a comparable low average concept relevance.

\subsection{Insights: class vs. anti-class concepts}
\label{sec:celeba}
The \celeba\;dataset with its 40 binary attributes represents an ideal test bed to investigate the behavior of concepts for (anti-)class (where a particular attribute is either present or absent) samples. For definiteness, we demonstrate this behavior for the ``male'' attribute and refer the reader to the supplementary material for further attributes. This experiment provides a perfect example for unique insights that can solely be gained from concept-based methods, which for example reveal the subtle difference between female and male nose concepts (see below). Importantly, these insights would remain hidden for standard local attribution methods (which would only highlight the nose region in both cases).

\begin{figure}
	\centering
	\def\arraystretch{0.9} 		
	\setlength\tabcolsep{5pt}		
		\begin{tabular}{lrlcr} 
			\multicolumn{2}{c}{male concepts} 	&	&			\multicolumn{2}{c}{female concepts} \\  \cline{0-1}   \cline{4-5}  
			male samples \rule{0mm}{3.5mm} 	& 		female samples   &		&	female samples		& 		male samples \\ 
			\multicolumn{2}{c}{\includegraphics[width=0.45\columnwidth]{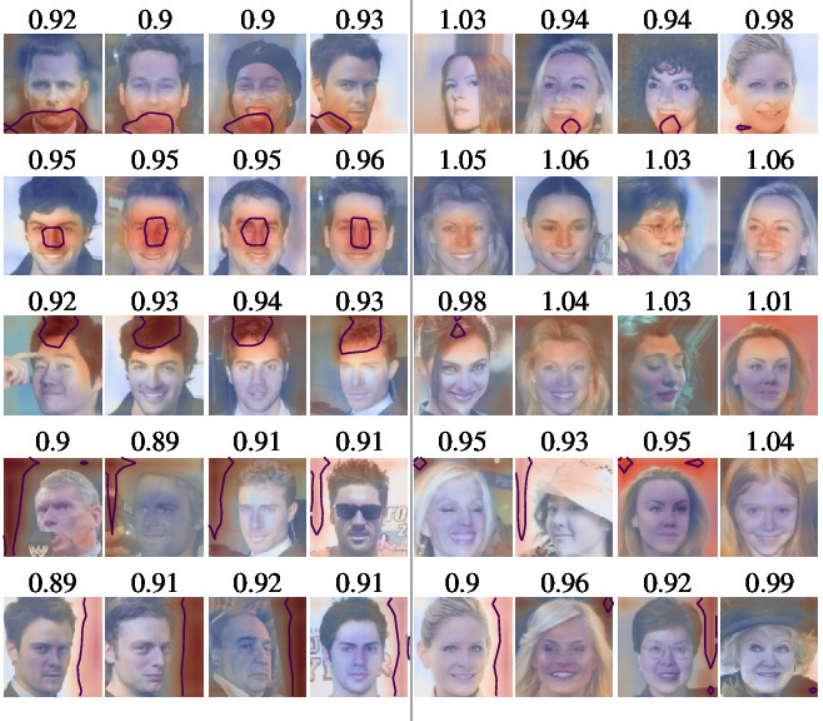}}	& &
			\multicolumn{2}{c}{\includegraphics[width=0.45\columnwidth]{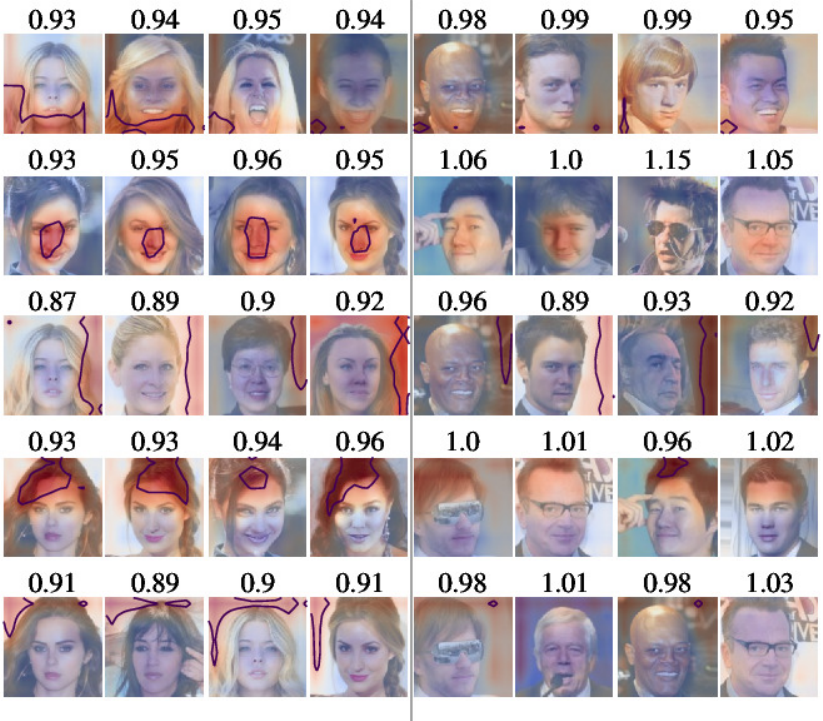}}	\\
		\end{tabular}
	
	\caption{SSCCD concepts on CelebA for a ResNet50 model for samples with the ``male'' (left) and ``female'' (right) attribute. Concepts are ordered according to average concept relevance showing samples with smallest \textit{sample-concept proximity}. We apply the same male concept on a random selection of female samples and vice-versa. Subtitles indicate the \textit{normalized sample-concept proximity} $\delta^{iA}/\delta^{A}$, i.e.\ samples with values $<1$ are (highly) activated. The ``male'' concepts can be roughly identified as chin, nose, hair(line), background, background and neck/hair, nose, background, hair, background for ``female''.}
	\label{fig:celeba_example}
\end{figure}

The left panel of \Cref{fig:celeba_example} shows the most important male concepts exploited by the model. 
These can be tagged as chin~(1.1), nose~(1.0), hair(line)~(0.5) as well as two (irrelevant) background concepts~($-$0.3, $-$0.8), where values in brackets denote the average concept relevance. This, in combination with similar findings for further attributes presented in the supplementary material, entails that the number of discriminative concepts is rather small, which seems plausible given the homogeneity of the dataset and the rather simple binary classification tasks.
Not surprisingly, the male concepts mostly activate for the corresponding regions in the female face, when evaluated on female samples.  
However, for the non-background concepts, concept proximities barely reach values below the respective concept activation threshold.
We quantify this statement by measuring their average \textit{normalized sample-concept proximity} scores across the whole dataset, $\delta^{\alpha A}/\delta^{A}$, where $\delta^{A}$ is the activation threshold. Values larger than one indicate that the concept is on average not activated.
We find (0.95, 0.99, 0.98) for male and (0.99, 1.05, 1.01) for female samples. Thus, only the male chin concept is activated on the female samples by a small margin whereas both nose and hair concepts remain unactivated. In the right panel of \Cref{fig:celeba_example}, we visualize female concepts, which interestingly show a slightly differently localized chin/hair concept as most important concept. To summarize, the identified concepts are highly specific both in terms of localization (c.f. chin for ``male'' vs.\ neck/hair for ``female'') and in terms of underlying features. The latter can be inferred from the fact that similarly localized concepts (e.g. nose for ``male''/``female'', hair for ``young''/``old'', see the supplementary material) are found for class and anti-class but evaluating a particular concept only weakly activates the corresponding region with the anti-concept.  As a final remark, we envision that SSCCD can be used as a detection tool for model biases (e.g. gender or race-related). This would entail comparing concepts discovered from a biased selection of samples to a balanced selection of samples.

\subsection{Insights: Concept similarities}
\label{sec:concept_similarity}
To demonstrate the soundness of the concept definition and distance measure along with the transferability of concepts across different classes, we examine concepts for a handpicked selection of four vehicle classes from \imagenet{} (``beach wagon'', ``pickup'', ``tow truck'', ``trailer truck''). We use a VGG16 model in this experiment.
Next, we extract four concepts for each of the classes with  SSCCD as well as for a vehicle super class, which is formed by the union of samples from all four classes.
This results in a total of 20 concepts, which we tagged based on visual inspection.
Within this setup, we can then systematically investigate concept commonalities and differences.

\begin{figure}
	\centering
	\includegraphics[width=0.4\linewidth]{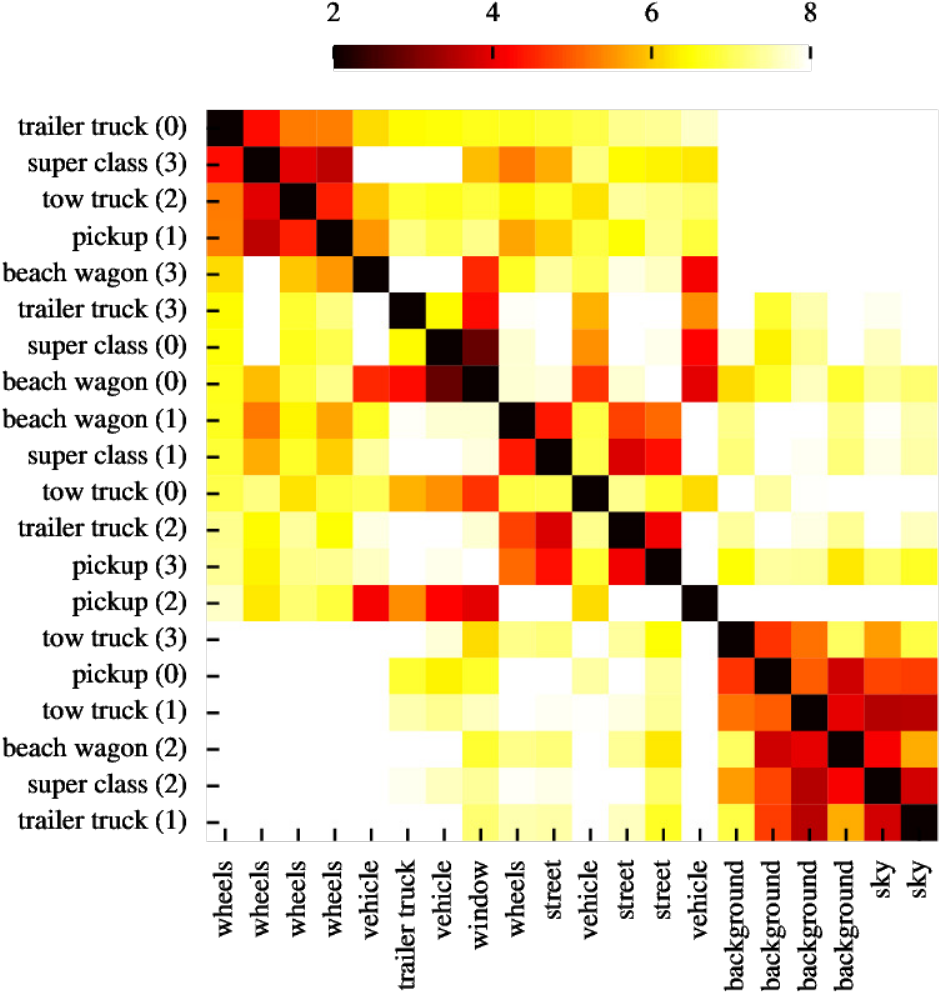}
	\includegraphics[width=0.48\linewidth]{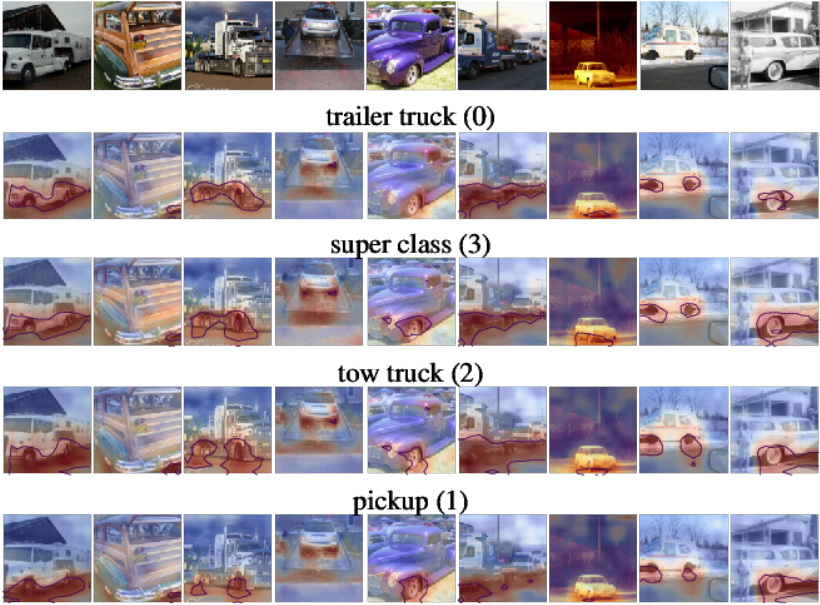}
	\caption{Left: Concept similarities (based on the Grassmann distance) between selected \imagenet{} vehicle concepts. 
		Rows and columns are sorted correspondingly (``trailer truck (0)''). 
		Rows are labeled according to class name and concept number whereas columns show manually tagged labels. 
		High concept similarities correspond to small Grassmann distances. 
		Note that background and object concepts are disentangled.
		 Right: Transferring and comparing the ``wheel'' concept from different vehicle classes on random test samples from different classes (first row).
		The rows below show concept maps obtained from applying the four most similar ``wheel'' concepts from \Cref{fig:concept_similarity}.
	}
	\label{fig:concept_similarity}
\end{figure}

We start with Grassmann distances between all concepts and show the resulting similarity matrix in \Cref{fig:concept_similarity}. Its rows/columns are sorted according to the trailer truck (0, wheel) concept similarity.
Clearly, semantically similar concepts are closely related and form a block-diagonal structure.
Most prominently tire and background concepts are blocked together. Importantly, both \emph{high-level} concepts are separated and not connected.
We stress that the manually tagged labels have been sorted solely based on a single Grassmann distance.
This further validates our approach based on principle angles between concept subspaces. 

Next, we perform a two-fold experiment. Firstly, akin to the previous analysis, we compare the four most similar concepts compared to the trailer truck (0)/'wheel' concept. Secondly, we test concept transferability by applying all concepts onto random test samples originating from different classes.
This is non-trivial and usually not possible with simple one-dimensional concept spaces.
Results are shown in \Cref{fig:concept_similarity}. All four-wheel concepts are visually nearly indistinguishable.
Additionally, all concepts transfer to samples from different classes. Again, this is fundamentally tied to the robust subspaces found by SSCCD.
These findings further hint towards actual regions in feature space, which are activated if a certain pattern/concept is present in the input. 
In particular, these regions are found independent whether trained on single- or on superclass training samples. 
Similar results for the ``window'' and ``street'' concepts are presented in the supplementary material and further support our arguments. 
Overall, our experiment strengthens the notion of robustness, transferability and consistency associated with SSCCD.

\subsection{Concept flipping}
To provide a quantitative comparison to previous concept discovery methods, we invoke the Smallest Destroying Concepts (SDC) benchmark as proposed in \cite{ace2019} and \cite{wu2020}. 
To evaluate SDC, we subsequently remove concepts, i.e. the corresponding segments, in the order of their sample-wise relevance starting from high to low.
We use images from ten \imagenet{} classes, which roughly align with CIFAR10 classes, and measure the (maximum) model accuracy on the selected subset of classes.
We use a ResNet50 model and extract concepts from the penultimate ResNet block for ACE \cite{ace2019}, the PCA/ICE-baseline described in \Cref{sec:mutlidim} and SSCCD.
ACE does not provide a measure of per-sample concept relevance.
Therefore, we revert to the order of their TCAV scores after discarding concepts where statistical testing in comparison to random input samples fails to stay below $p=0.05$.
We remove segments based on a classical imputation algorithm \cite{bertalmio2001navier}.
This results in more realistic imputed images. Thus, the model is evaluated \emph{on-manifold} in contrast to imputing with gray patches as often done in the literature.
For similar reasons, we avoid the Smallest Sufficient Concepts (SSC) benchmark, which would require high-quality imputation algorithms to avoid evaluating the model far from the data manifold.
For SSCCD and the PCA/ICE-baseline, we compute concept relevance by employing \pd with the same imputation method in input space.
It is worth stressing again that SSCCD is indifferent to attribution methods. In particular, the concept ranking is not sensitive to this choice.

\begin{figure}
	\centering
		\def\arraystretch{0.9} 		
	\setlength\tabcolsep{0.3pt}		
		\begin{tabular}{cccc} 
		\rule{9.mm}{0.mm}SSCCD \rule{8mm}{0.mm} 	& \rule{9.5mm}{0.mm}PCA \rule{10mm}{0.mm} &ACE   &	\rule{6mm}{0.mm} SDC experiment	\\
		\multicolumn{3}{c}{\includegraphics[width=0.7\columnwidth]{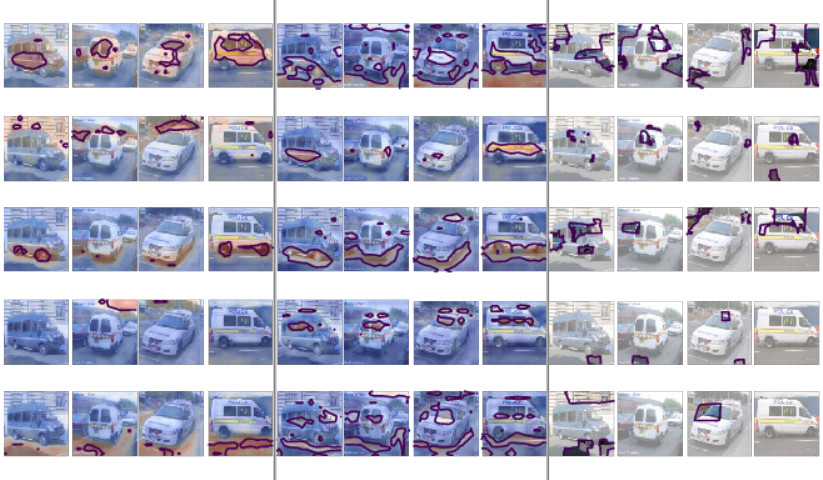}}	&
			\multicolumn{1}{c}{\includegraphics[width=0.28\columnwidth]{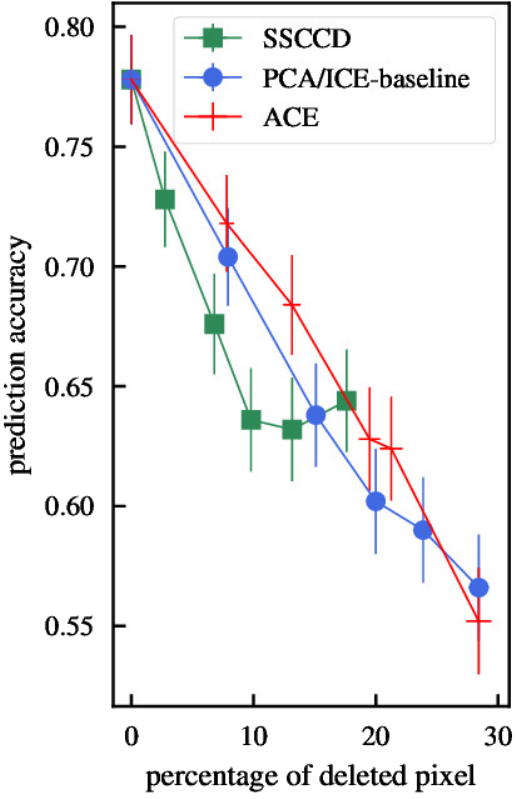}}	\\
			\end{tabular}		
	\caption{Left: Qualitative comparison between SSCCD, PCA/ICE-baseline and ACE based on random validation set samples from the \imagenet{} police van class. Right: Concept flipping (Smallest Destroying Concepts) benchmark comparing SSCCD and ACE\cite{ace2019}.
		Concepts are flipped once at a time in descending order of concept relevance/TCAV score, respectively. 
		We measure the decline in model accuracy. 
		Meaningful concept discovery and quantification methods are supposed to show a sharp decline in this figure. 
		The importance of the attribution method is assessed by also flipping discovered concept segments in random order.
	}	\label{fig:sdc_benchmark}
\end{figure}

In the left panel of \Cref{fig:sdc_benchmark}, we qualitatively compare concepts identified by ACE, ICE/PCA and SSCCD. The ACE concepts are expressed only in a fraction of the images and, opposed to SCCDD, it is difficult to identify coherent, meaningful structures. This is in contrast to the much more coherent impression of the same ACE concepts visualized on training set samples, see supplementary material. Therefore, we suspect that generalization of concepts identified on training set samples to the test set is difficult for ACE. As observed in \Cref{sec:mutlidim}, the PCA concepts in \Cref{fig:sdc_benchmark} are less coherent and localized than SSCCD concepts.

In the right panel of \Cref{fig:sdc_benchmark}, we show the results of the SDC experiment. 
In contrast to previous studies \cite{ace2019,wu2020}, we report the model performance depending on the fraction of occluded pixels, which is essential for comparability since the segment size varies between different approaches.
SSCCD shows the sharpest decline and is hence considered the superior approach for concept-based explanations. The slight increase in accuracy after flipping the last SSCCD concept is likely due to flipping i.e. removing slightly negatively attributed features from the background while parts of the relevant region are still visible (as previous concepts are localized).
While SDC as a benchmark for concept explanations shows a favorable outcome for the proposed method, it entails several shortcomings:
The quantitative results, albeit in most cases not the method ranking, may strongly depend on the imputation algorithm.
Model scores for heavily occluded images are potentially unreliable due to off-manifold evaluation.
Lastly, one needs to disentangle concept attribution and discovery. This ideally allows to change only a single component and therefore enable for more consistent comparisons.

\section{Summary and Discussion}
In this work, we provide a new definition of concepts as linear subspaces in feature space and present a constructive way of discovering them by means of sparse subspace clustering. 
Our experiments demonstrate a number of insights from SSCCD that are infeasible to obtain from standard pixel-wise attributions or existing concept-based interpretability methods.
These range from a similarity measure for concept subspaces,
over to \textit{concept maps} that spatially resolve the concept expression for a specific sample, 
to the ability to quantify a concept's relevance on a per-example basis via a user-specified local attribution method.

\section*{Acknowledgements}
This work was funded by the German Ministry for Education and Research as BIFOLD - Berlin Institute for the Foundations of Learning and Data (ref. 01IS18025A and ref. 01IS18037A).

\bibliography{ssccd}

\appendix
\section{One-dimensional SSCCD concepts (Section 4.1)}
\begin{figure}[ht]
	\centering
		\def\arraystretch{0.5} 		
	\setlength\tabcolsep{0pt}		
	\begin{tabularx}{\textwidth}{*{3}{>{\centering\arraybackslash\hsize=0.3334\textwidth}X}}
		SSCCD	&	SSCCD 1D 	& PCA \\
		\multicolumn{3}{c}{	\includegraphics[width=\columnwidth]{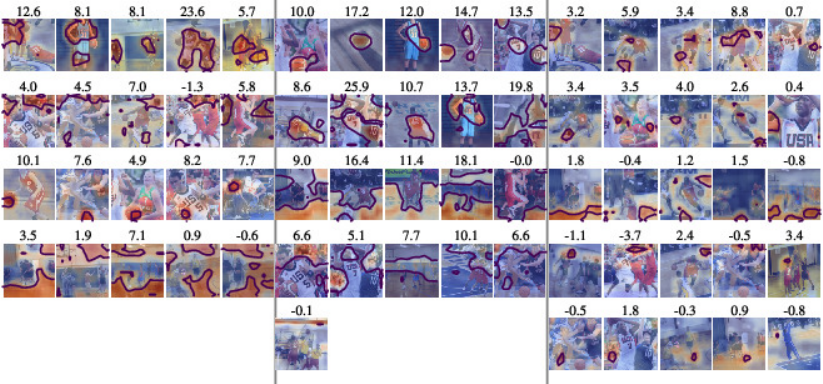}} \\
	\end{tabularx}
	\caption{Concept maps for a VGG16 model (final convolutional layer) and the ``basketball'' class with high intra-class diversity. We compare multi-dimensional SSCCD concepts (left) to SSCCD concepts with a one-dimensional basis (middle) and one-dimensional concepts (right) obtained via direct application of PCA in feature space. In all three cases rows are ordered according to average concept relevance.}
	\label{fig:n_concepts_basketball}
\end{figure}

\begin{figure}[ht]
	\centering
			\def\arraystretch{0.5} 		
	\setlength\tabcolsep{0pt}		
	\begin{tabularx}{\textwidth}{*{3}{>{\centering\arraybackslash\hsize=0.3334\textwidth}X}}
		SSCCD	&	SSCCD 1D 	& PCA \\
		\multicolumn{3}{c}{	\includegraphics[width=\columnwidth]{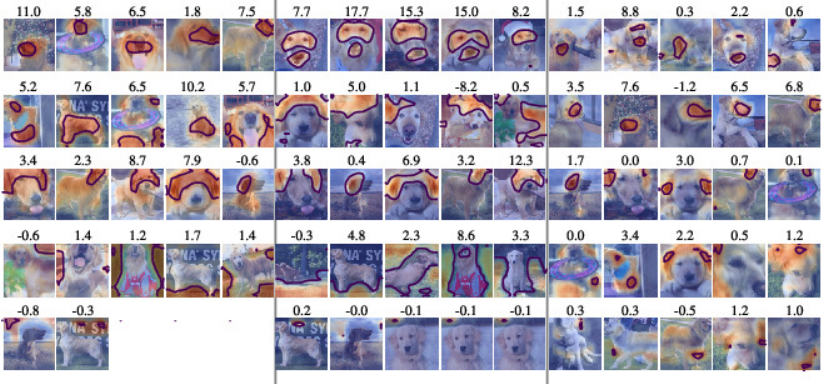}} \\
	\end{tabularx}
	\caption{Concept maps for a VGG16 model (final convolutional layer) and the ``golden retriever'' with low intra-class diversity. We compare multi-dimensional SSCCD concepts (left) to SSCCD concepts with a one-dimensional basis (middle) and one-dimensional concepts (right) obtained via direct application of PCA in feature space. In all three cases rows are ordered according to average concept relevance.}
	\label{fig:n_concepts_goldenretriever}
\end{figure}

\begin{figure}[ht]
	\centering
				\def\arraystretch{0.5} 		
	\setlength\tabcolsep{0pt}		
	\begin{tabularx}{\textwidth}{*{3}{>{\centering\arraybackslash\hsize=0.3334\textwidth}X}}
		SSCCD	&	SSCCD 1D 	& PCA \\
		\multicolumn{3}{c}{	\includegraphics[width=\columnwidth]{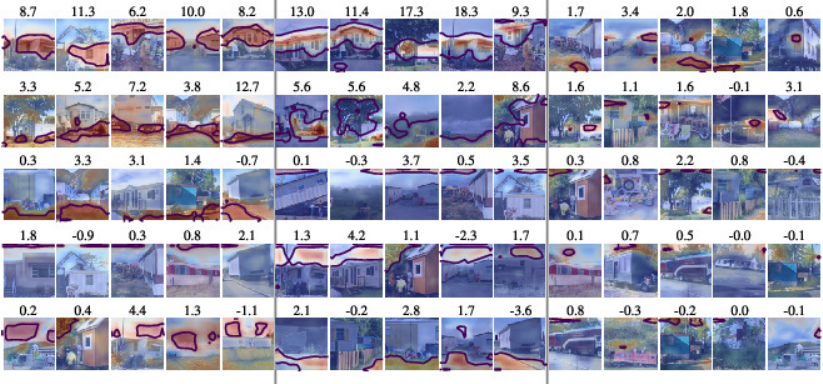}} \\
	\end{tabularx}
	\caption{Concept maps for a VGG16 model (final convolutional layer) and the ``mobile home'' class with high intra-class diversity. We compare multi-dimensional SSCCD concepts (left) to SSCCD concepts with a one-dimensional basis (middle) and one-dimensional concepts (right) obtained via direct application of PCA in feature space. In all three cases rows are ordered according to average concept relevance.}
	\label{fig:n_concepts_mobilehome}
\end{figure}

\begin{figure}[ht]
	\centering
	\def\arraystretch{0.5} 		
	\setlength\tabcolsep{0pt}		
	\begin{tabularx}{\textwidth}{*{3}{>{\centering\arraybackslash\hsize=0.3334\textwidth}X}}
		SSCCD	&	SSCCD 1D 	& PCA \\
		\multicolumn{3}{c}{	\includegraphics[width=\columnwidth]{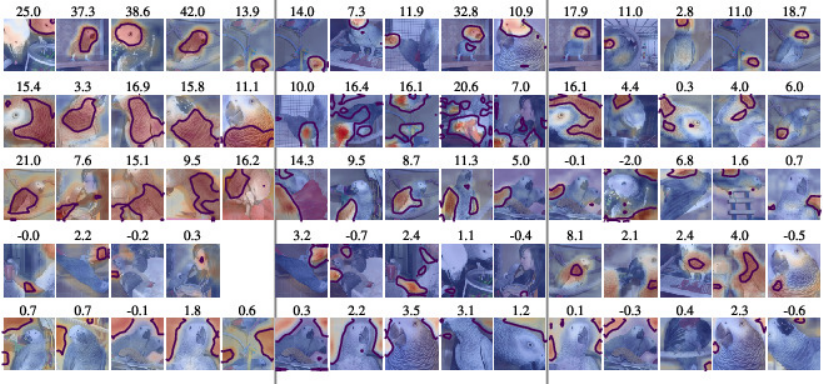}} \\
	\end{tabularx}
	\caption{Concept maps for a VGG16 model (final convolutional layer) and the ``african grey'' class with low intra-class diversity. We compare multi-dimensional SSCCD concepts (left) to SSCCD concepts with a one-dimensional basis (middle) and one-dimensional concepts (right) obtained via direct application of PCA in feature space. In all three cases rows are ordered according to average concept relevance.}
	\label{fig:n_concepts_africangrey}
\end{figure}

\begin{figure}[ht]
	\centering
		\def\arraystretch{0.5} 		
	\setlength\tabcolsep{0pt}		
	\begin{tabularx}{\textwidth}{*{3}{>{\centering\arraybackslash\hsize=0.3334\textwidth}X}}
		SSCCD	&	SSCCD 1D 	& PCA \\
		\multicolumn{3}{c}{	\includegraphics[width=\columnwidth]{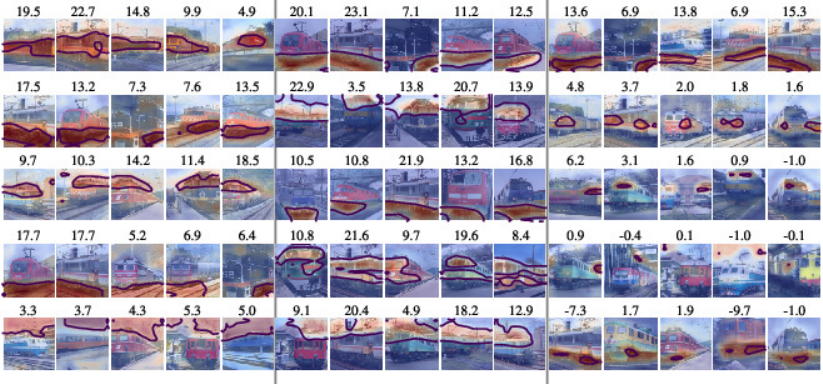}} \\
	\end{tabularx}
	\caption{Concept maps for a VGG16 model (final convolutional layer) and the ``electric locomotive'' class with high intra-class diversity. We compare multi-dimensional SSCCD concepts (left) to SSCCD concepts with a one-dimensional basis (middle) and one-dimensional concepts (right) obtained via direct application of PCA in feature space. In all three cases rows are ordered according to average concept relevance.}
	\label{fig:n_concepts_electriclocomotive}
\end{figure}

\begin{figure}[ht]
	\centering
		\def\arraystretch{0.5} 		
	\setlength\tabcolsep{0pt}		
	\begin{tabularx}{\textwidth}{*{3}{>{\centering\arraybackslash\hsize=0.3334\textwidth}X}}
		SSCCD	&	SSCCD 1D 	& PCA \\
		\multicolumn{3}{c}{	\includegraphics[width=\columnwidth]{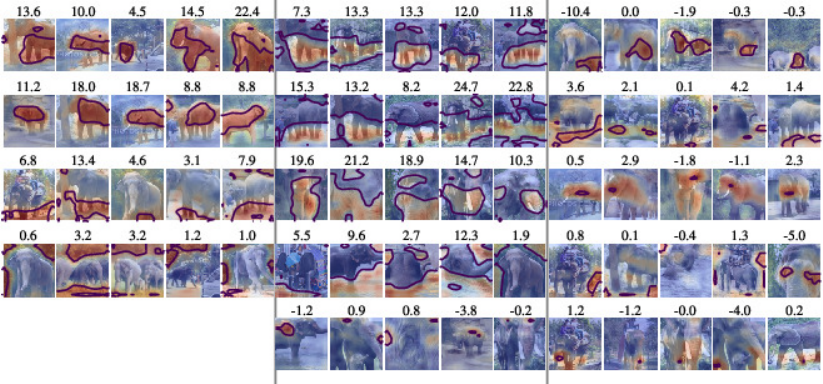}} \\
	\end{tabularx}
	\caption{Concept maps for a VGG16 model (final convolutional layer) and the ``indian elephant'' class  with low intra-class diversity. We compare multi-dimensional SSCCD concepts (left) to SSCCD concepts with a one-dimensional basis (middle) and one-dimensional concepts (right) obtained via direct application of PCA in feature space. In all three cases rows are ordered according to average concept relevance.}
	\label{fig:n_concepts_indianelephant}
\end{figure}

In \Cref{fig:n_concepts_basketball} to \Cref{fig:n_concepts_indianelephant}, we revisit the discussion on the multidimensional nature of concepts. We use the ``basketball'', ``mobile home'' and ``electric locomotive'' class as representatives for a class with a high intra-class variance and the ``golden retriever'', ``african grey'' and ``indian elphant'' class as representatives for a class with small intra-class variance. As discussed in the main text, the PCA concepts come out more consistent in the latter case.  We go beyond the discussion in the main text by also showing concept maps, where we reduce the subspaces identified via SSCCD to their first PCA component i.e.\ one-dimensional subspaces. This can be used to identify the dominant aspect underlying a specific concept subspace, it avoids the issues of double-counting closely related directions in feature space that would have appeared as separate concepts in the PCA approach. These results indicate that also one-dimensional concepts can entail a certain value but only if they are determined in an appropriate way. In this case, they are based on the already identified multidimensional SSCCD concepts, which could in this case be used directly instead.

\section{More ImageNet classes for VGG16/ResNet50 (Section 4.2)}
\begin{figure}[ht]
	\centering
		\def\arraystretch{1} 		
		\setlength\tabcolsep{0pt}		
		\begin{tabularx}{\textwidth}{*{4}{>{\centering\arraybackslash\hsize=0.25\textwidth}X}}
			\multicolumn{2}{c}{VGG16} & \multicolumn{2}{c}{ResNet50} \\ 
			\scriptsize{highly activated}	 & \scriptsize{randomly selected} &\scriptsize{highly activated} & \scriptsize{randomly selected}\\
			\multicolumn{4}{c}{\includegraphics[width=\columnwidth]{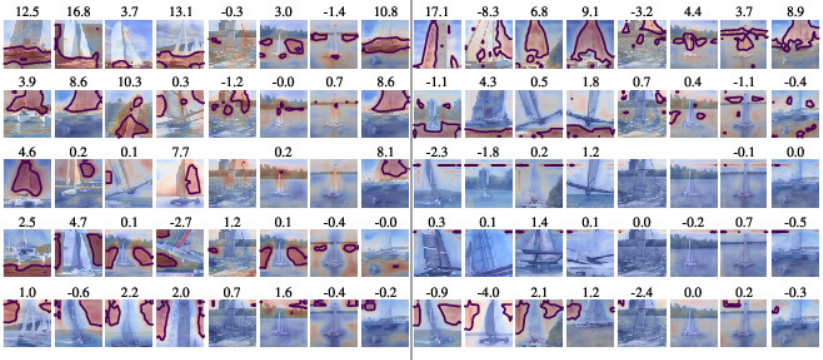}} \\
		\end{tabularx}
	\caption{Concept maps for the ImageNet class ``trimaran''. For VGG16 the concepts can be roughly identified with hull, mast/sail, rigging, water, sky and for ResNet50 with rigging, water, empty/sky, empty/sky, sky.}
	\label{fig:arch_compare1}
\end{figure}

\begin{figure}[ht]
	\centering
	\def\arraystretch{1} 		
	\setlength\tabcolsep{0pt}		
	\begin{tabularx}{\textwidth}{*{4}{>{\centering\arraybackslash\hsize=0.25\textwidth}X}}
		\multicolumn{2}{c}{VGG16} & \multicolumn{2}{c}{ResNet50} \\ 
		\scriptsize{highly activated}	 & \scriptsize{randomly selected} &\scriptsize{highly activated} & \scriptsize{randomly selected}\\
		\multicolumn{4}{c}{	\includegraphics[width=\columnwidth]{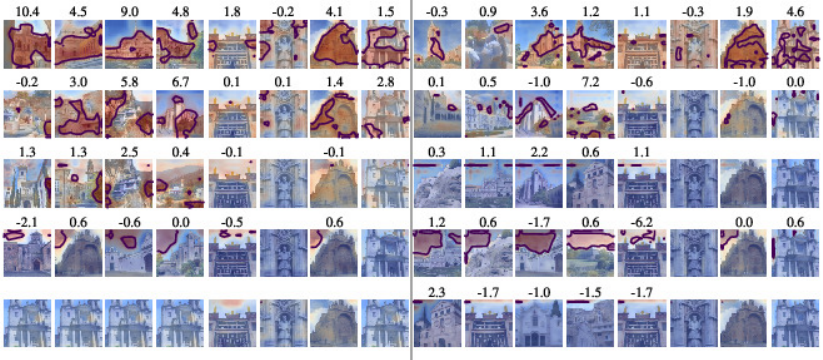}} \\
	\end{tabularx}
		\caption{Concept maps for the ImageNet class ``monastery''. For VGG16 the the concepts can be roughly identified with fancy facade, simple facade, trees, sky, empty and for ResNet50 with facade, roof, empty/sky, sky, empty/sky.
		}
	\label{fig:arch_compare2}
\end{figure}

\begin{figure}[ht]
	\centering
		\def\arraystretch{1} 		
	\setlength\tabcolsep{2pt}		
	\begin{tabularx}{\textwidth}{*{4}{>{\centering\arraybackslash\hsize=0.23\textwidth}X}}
		\multicolumn{2}{c}{VGG16} & \multicolumn{2}{c}{ResNet50} \\
				\scriptsize{highly activated}	 & \scriptsize{randomly selected} &\scriptsize{highly activated} & \scriptsize{randomly selected}\\ 
		\multicolumn{4}{c}{	\includegraphics[width=\columnwidth]{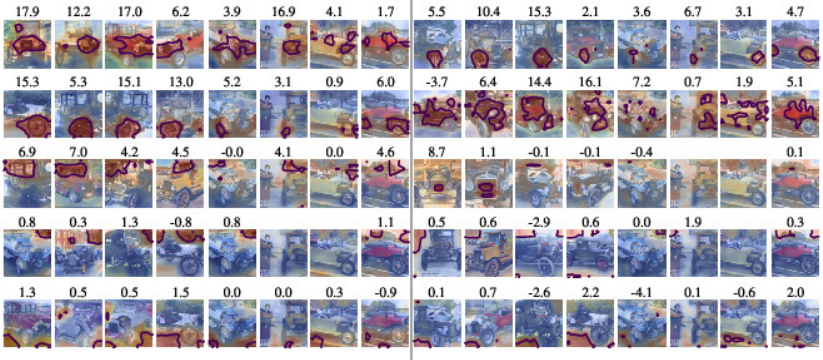}} \\

	\end{tabularx}

	\caption{Concept maps for the ImageNet class ``Model T''. For VGG16 the concepts can be roughly identified as lights, wheels, windows, trees, street and for ResNet50 as wheels, hood, radiator grill, sky, ground.}
	\label{fig:arch_compare3}
	\end{figure}

\begin{figure}[ht]
    \centering
    	\def\arraystretch{1} 		
    \setlength\tabcolsep{0pt}		
    \begin{tabularx}{\textwidth}{*{4}{>{\centering\arraybackslash\hsize=0.25\textwidth}X}}
    	\multicolumn{2}{c}{VGG16} & \multicolumn{2}{c}{ResNet50} \\ 
 	    	\scriptsize{highly activated}	 & \scriptsize{randomly selected} &\scriptsize{highly activated} & \scriptsize{randomly selected}\\ 
    	\multicolumn{4}{c}{	\includegraphics[width=\columnwidth]{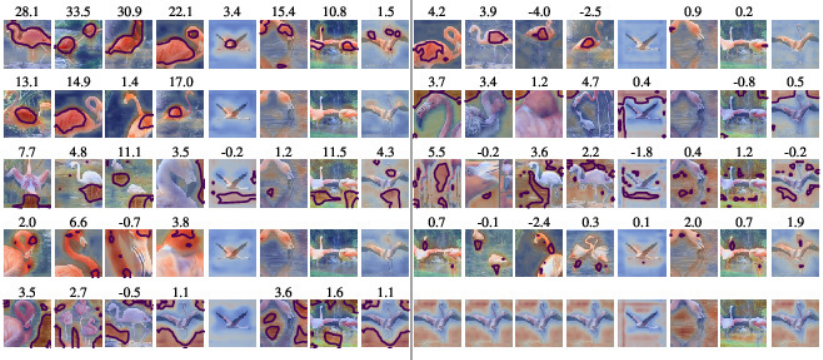}} \\
    \end{tabularx}
	\caption{Concept maps for the ImageNet class ``flamingo''.  For VGG16 the concepts can be roughly identified as body, wing feathers, legs, head/neck, water
	and for ResNet50 as feather wings, background, water, legs/neck, empty.}
	\label{fig:arch_compare4}
\end{figure}

In \Cref{fig:arch_compare1} to \Cref{fig:arch_compare4}, we present further  concept maps for four randomly selected ImageNet classes comparing VGG16 (last convolutional layer) to ResNet50 (penultimate ResNet block) concepts. 
All four figures are structured identically: 
Each row corresponds to a particular concept identified by SSCCD. 
The rows are sorted according to mean concept relevance (for every architecture separately) assessed via \ig{} in the input space. 
The eight leftmost pictures show VGG16 concept maps, the eight rightmost pictures correspond to ResNet50. 
Among these groups of eight pictures the four leftmost pictures show the pictures that were activated most by the concept (in terms of \textit{sample-concept proximity}), which gives in most cases the best overall impression what the concept signifies. 
The remaining four pictures show concept maps for four randomly selected pictures from the corresponding test dataset. These give an impression how consistently a concept is expressed in the dataset in terms of a selection with no selection bias. 
On top of each picture we also indicate the per-example concept relevance (again assessed via \ig{} in input space) for samples where the \textit{sample-concept proximity} reached a value below the concept-specific \textit{threshold angle}.

In most cases, one finds consistent human-identifiable concepts (with the exception of a few concepts that are not activated in the test set at all and a few elongated structures in the case of ResNet50 both of which we mark as ``empty'') which we also indicate in the figure captions. These reveal interesting differences between the model architectures that are supposed to be investigated in future studies.


\section{Layer-dependence (Section 4.2)}
\begin{figure}[ht]
	\centering
	ResNet50 block \\
  	penultimate \hspace{2.5cm} final \\
	\includegraphics[width=0.25\columnwidth]{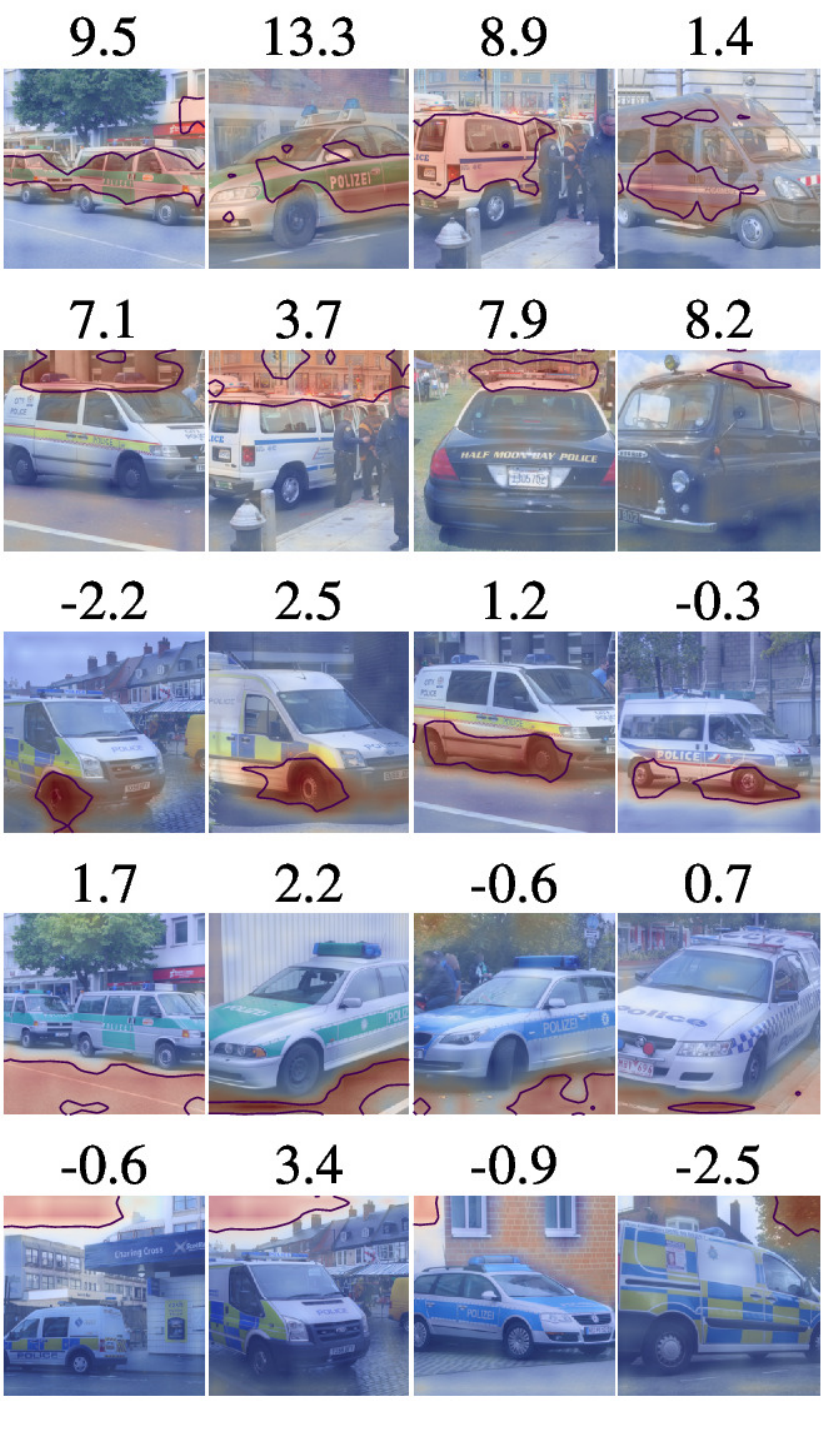}
	\includegraphics[width=0.25\columnwidth]{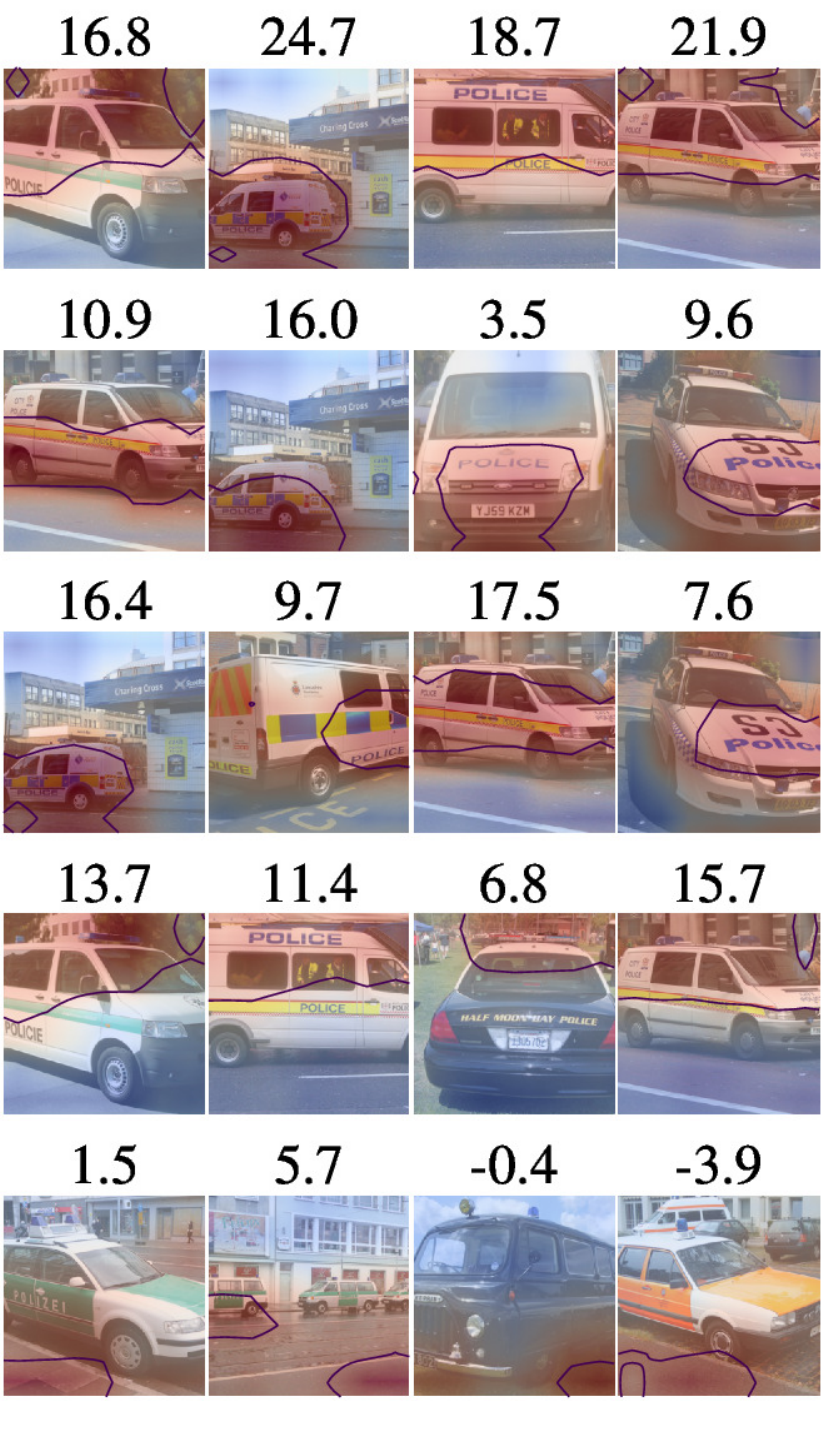}
	\caption{Concept maps for the ImageNet ``police van'' class at the penultimate ResNet50 block (left) and at the final block of a ResNet50 model.}
	\label{fig:layerdependence}
\end{figure}

We compare concepts of the penultimate ResNet50 block with concepts of the final block in \Cref{fig:layerdependence} to assess the dependency of SSCCD concepts on the model layer.
Concepts from the final block are broader than concepts at the penultimate block in the sense that they cover multiple components of the object, e.g. the first concept relates to the entire upper part of the police van including part of the car body, windows and buildings in the background. In contrast, concepts at the penultimate block can be identified with specific parts of the police van, like car body, emergency lights or tires. While this finding might be partially explained by the resizing factor between concept maps in feature and input space which is two times larger for the final block, it aligns with the common observation that the abstraction level increases for higher convolutional layers.

\section{Varying the number of concepts (Section 4.2)}
\begin{figure}[ht]
	\centering
					\def\arraystretch{0.01} 		
	\setlength\tabcolsep{5pt}		
	\begin{tabularx}{\columnwidth}{c c c}  

\includegraphics[width=0.31\columnwidth]{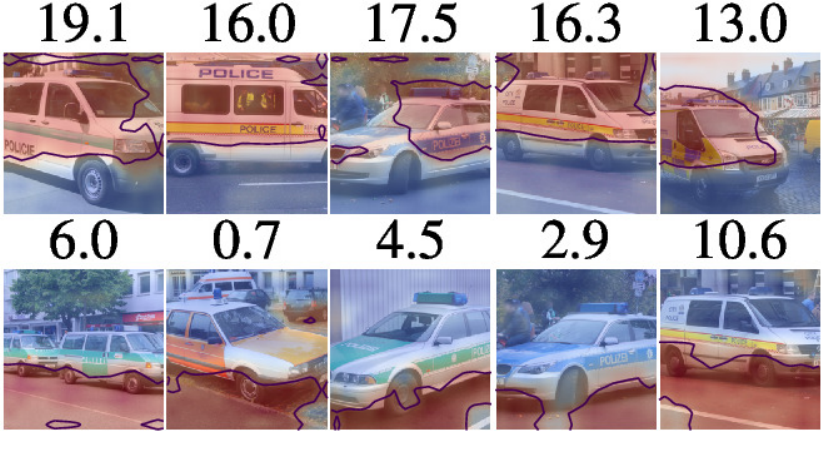}	&
		\includegraphics[width=0.31\columnwidth]{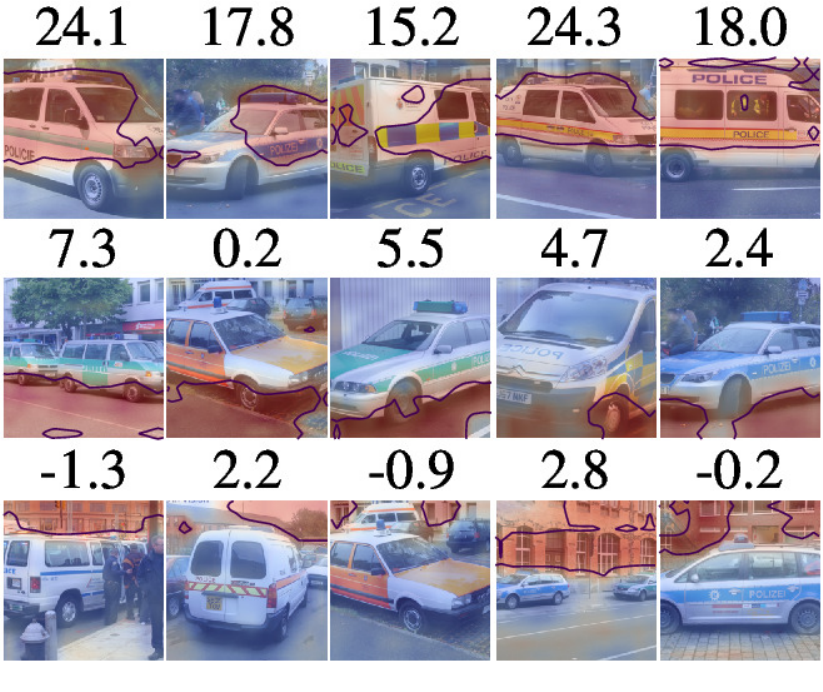}	&
		\includegraphics[width=0.31\columnwidth]{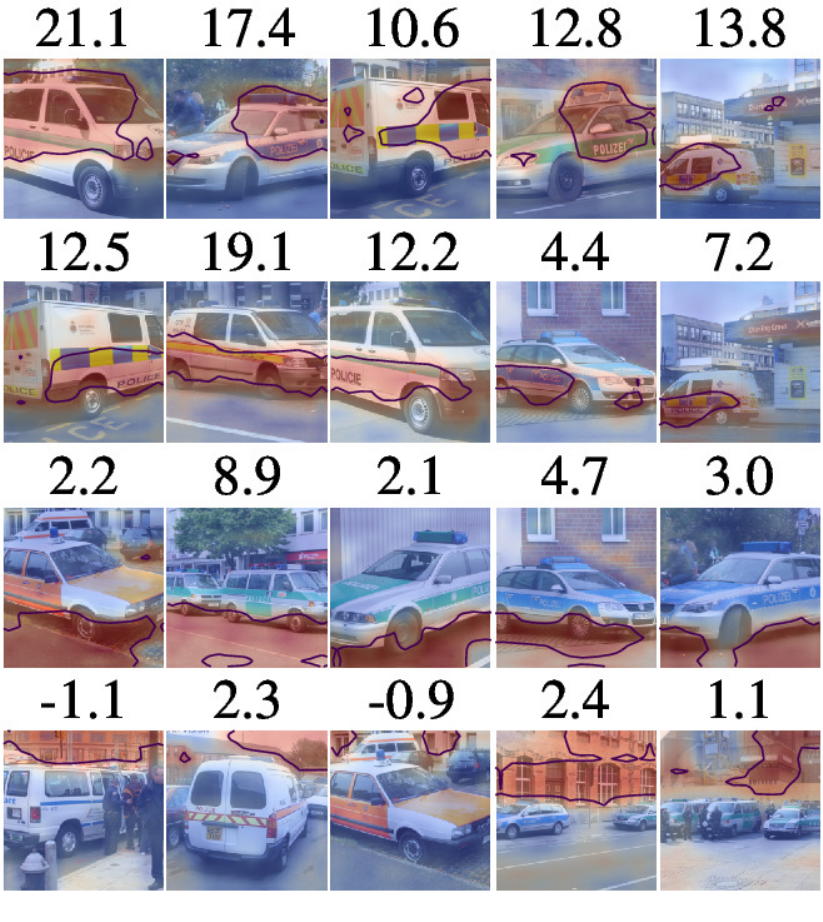}		\\
		 		$n_\text{concepts} = 2$ & $n_\text{concepts} = 3$  & $n_\text{concepts} = 4$  \\ 
		 		
		 	\rule{0mm}{6mm}	&& \\ 

		\includegraphics[width=0.31\columnwidth]{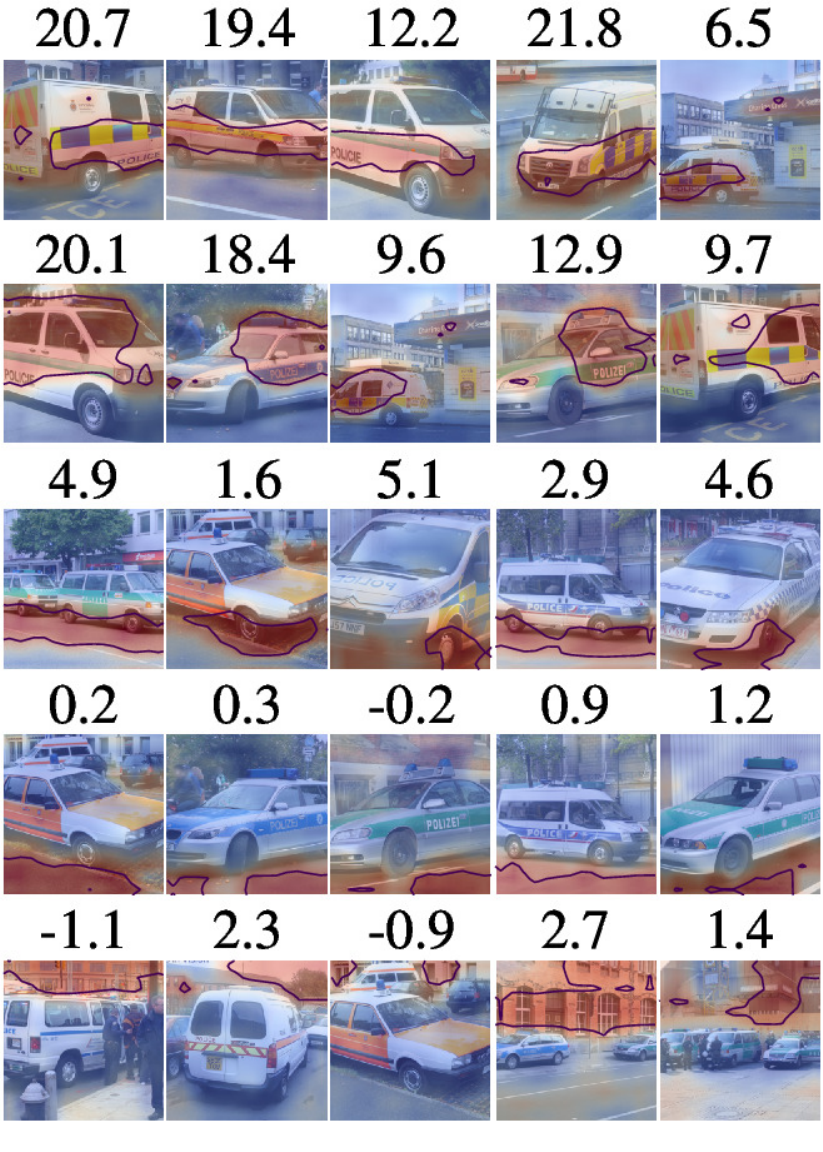}	&
		\includegraphics[width=0.31\columnwidth]{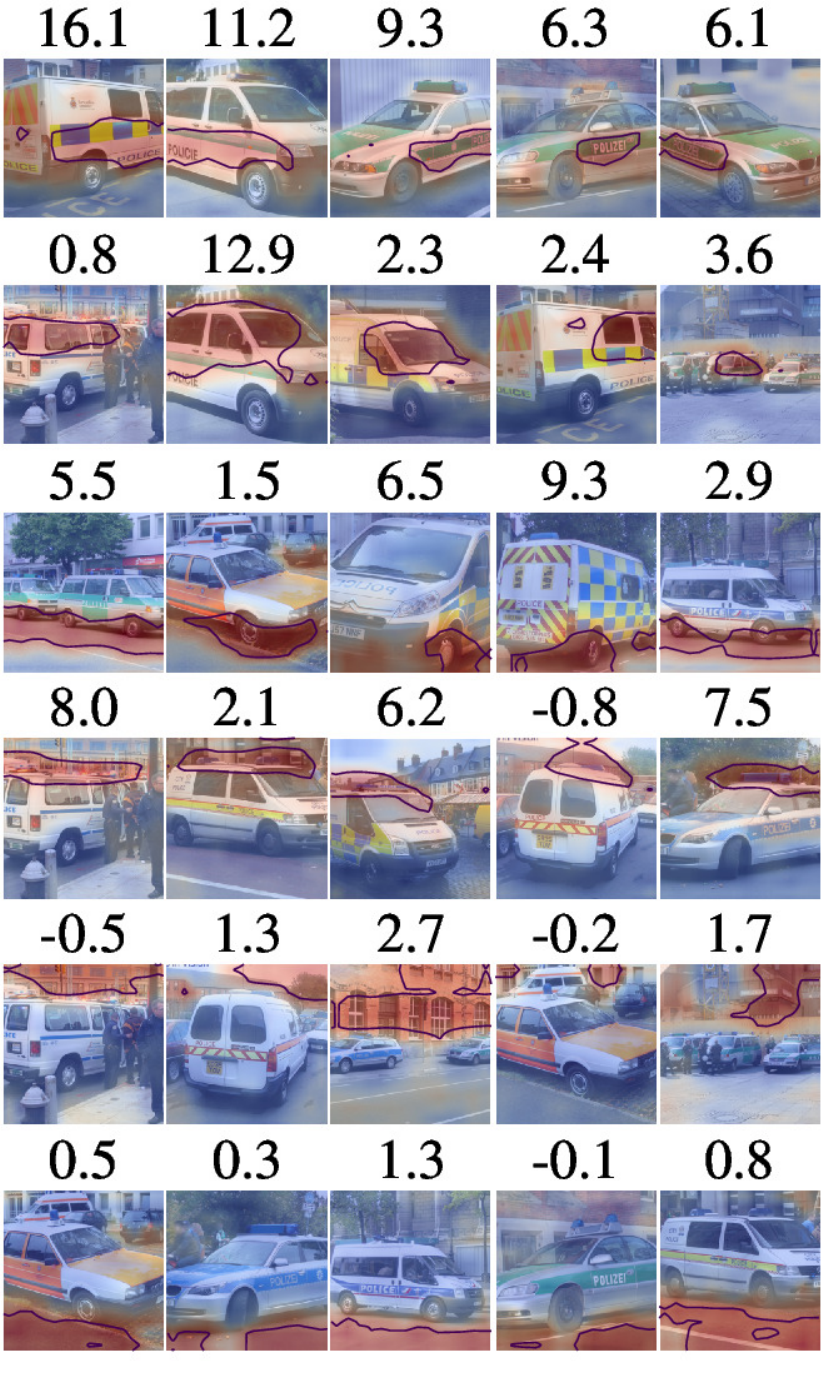}&
		\includegraphics[width=0.31\columnwidth]{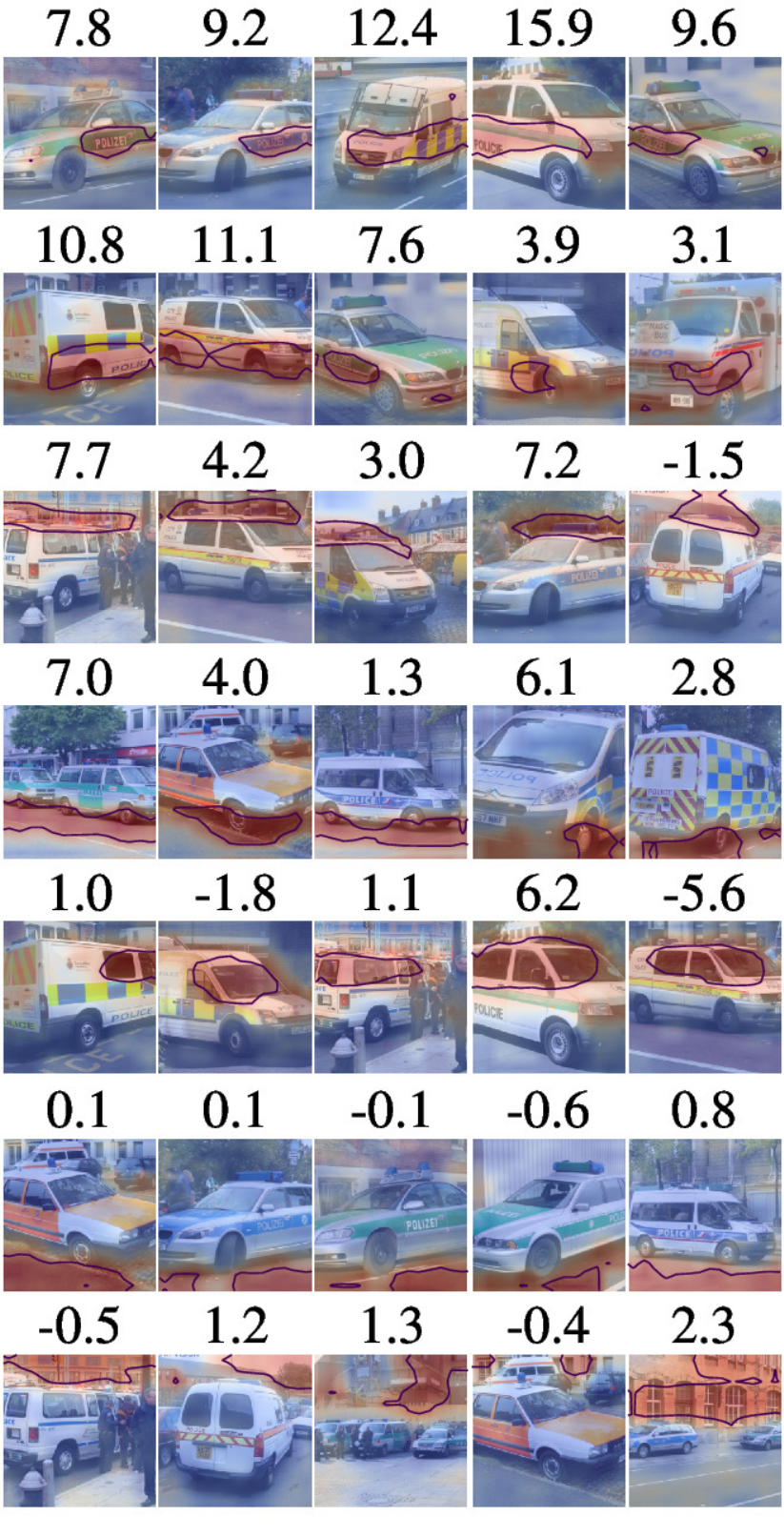} \\
						$n_\text{concepts} = 5$ & $n_\text{concepts} = 6$  & $n_\text{concepts} = 7$  
	\end{tabularx}

	\caption{Concept maps for police van class and a VGG16 model (final convolutional layer) varying the number of concepts from 2 to 7, where we show the most activated examples (in the sense of sample-concept proximity). Determining the number of concepts via the largest gap in the spectrum yields only yields a value of 2 in this case.}
	\label{fig:n_concepts}
\end{figure}

In \Cref{fig:n_concepts}, we demonstrate the impact of varying the number of concepts (for fixed, precomputed self-representation) at the example of the ImageNet class ``police van''. It shows how new concepts emerge and how existing concepts subdivide to form more finegrained concepts. For more than 5 concepts, also VGG16 also shows the ``emergency lights'' concept that was only observed for ResNet50 at a fixed number of 5 clusters.

\section{More CelebA (anti-)class concepts (Section 4.3)}
\begin{figure}[ht]
	\centering
				\def\arraystretch{0.9} 		
	\setlength\tabcolsep{5pt}		
		\begin{tabular}{lrlcr} \toprule
			\multicolumn{2}{c}{male concepts} 	&	&			\multicolumn{2}{c}{female concepts} \\  \cline{0-1}   \cline{4-5}  
			male samples \rule{0mm}{3.5mm} 	& 		female samples   &		&	female samples		& 		male samples \\ 
			\multicolumn{2}{c}{\includegraphics[width=0.45\columnwidth]{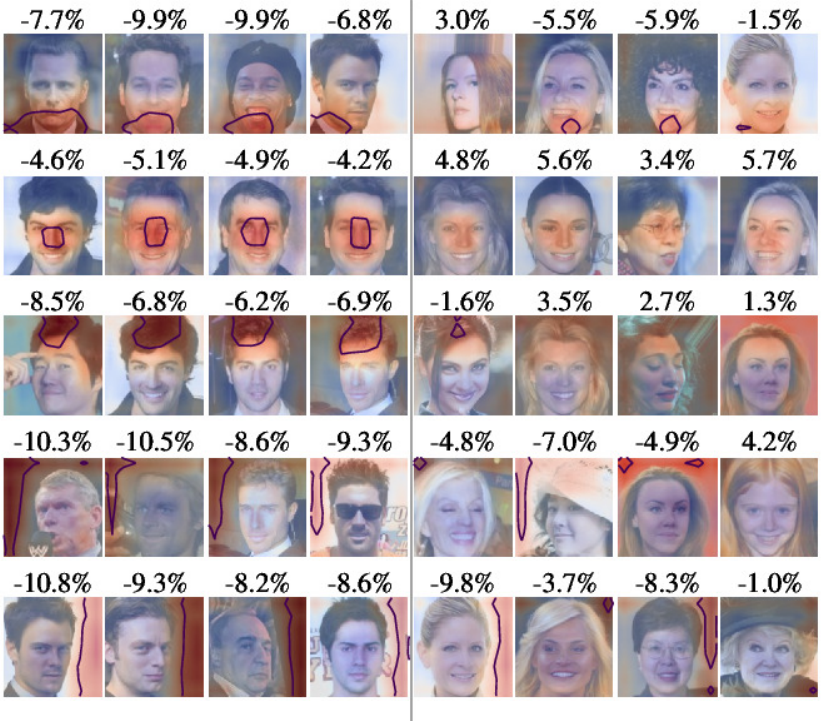}}	& &
			\multicolumn{2}{c}{\includegraphics[width=0.45\columnwidth]{celeba_male_5_True.pdf}}	\\
		\end{tabular}
	
	\caption{SSCCD concepts on CelebA for a ResNet50 model for samples with the ``male'' (left) and ``female'' (right) attribute. Concepts are ordered according to average concept relevance showing samples with smallest \textit{sample-concept proximity}. We apply the same male concept on a random selection of female samples and vice-versa. Subtitles indicate the \textit{sample-concept proximity} of samples relative to the \textit{normalized sample-concept proximity} $\delta^{iA}/\delta^{A}$, i.e.\ samples with values $<1$ are (highly) activated. The ``male'' concepts can be roughly identified as chin, nose, hair(line), background, background and neck/hair, nose, background, hair, background for ``female''.}	
	\label{fig:celeba_male}
\end{figure}

\begin{figure}[ht]
	\centering

			\def\arraystretch{0.9} 		
	\setlength\tabcolsep{5pt}		
		\begin{tabular}{lrlcr} \toprule
			\multicolumn{2}{c}{young concepts} 	&	&			\multicolumn{2}{c}{old concepts} \\  \cline{0-1}   \cline{4-5}  
			young samples \rule{0mm}{3.5mm} 	& 		old samples   &		&	old samples		& 		young samples \\ 
			\multicolumn{2}{c}{\includegraphics[width=0.45\columnwidth]{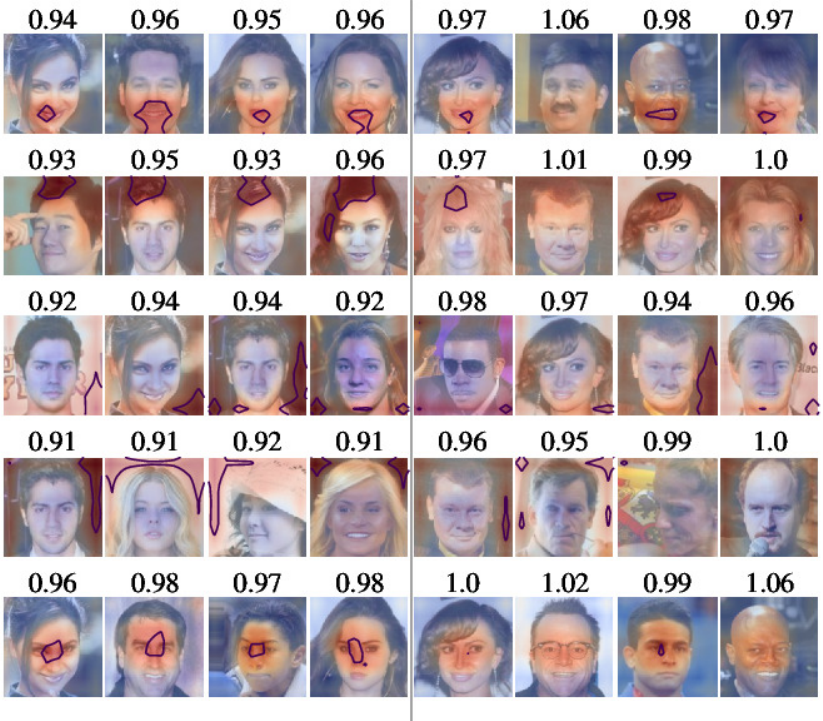}}	& &
			\multicolumn{2}{c}{\includegraphics[width=0.45\columnwidth]{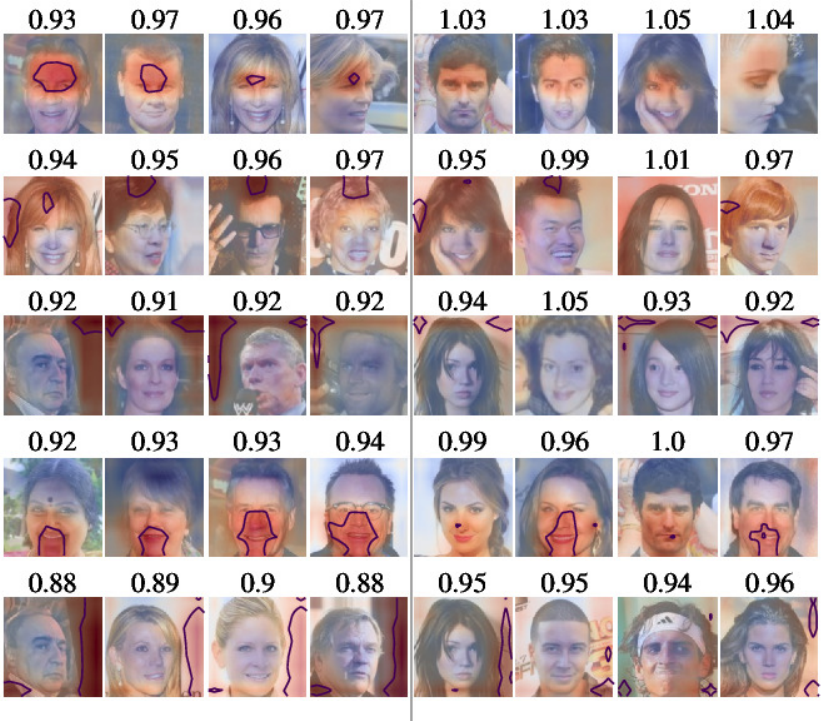}}	\\
		\end{tabular}
	
	\caption{SSCCD concepts on CelebA for a ResNet50 model for samples with the ``young'' (left) and ``old'' (right) attributes. Concepts are ordered according to average concept relevance showing samples with smallest \textit{sample-concept proximity}. We apply the same young concept on a random selection of old samples and vice-versa. Subtitles indicate the \textit{sample-concept proximity} of samples relative to the \textit{normalized sample-concept proximity} $\delta^{iA}/\delta^{A}$, i.e.\ samples with values $<1$ are (highly) activated. The ``young'' concepts can be roughly identified with mouth, hair, background, background, between eyes and forehead, hair, background, mouth/nose, background for ``old''.}	
	\label{fig:celeba_young}
\end{figure}

\begin{figure}[ht]
	\centering
		\def\arraystretch{0.9} 		
	\setlength\tabcolsep{5pt}		
			\begin{tabular}{lrlcr} \toprule
		\multicolumn{2}{c}{smiling concepts} 	&	&			\multicolumn{2}{c}{not smiling concepts} \\  \cline{0-1}   \cline{4-5}  
		smiling samples \rule{0mm}{3.5mm} 	& 		not smiling samples   &		&	not smiling samples		& 		smiling samples \\ 
		\multicolumn{2}{c}{\includegraphics[width=0.45\columnwidth]{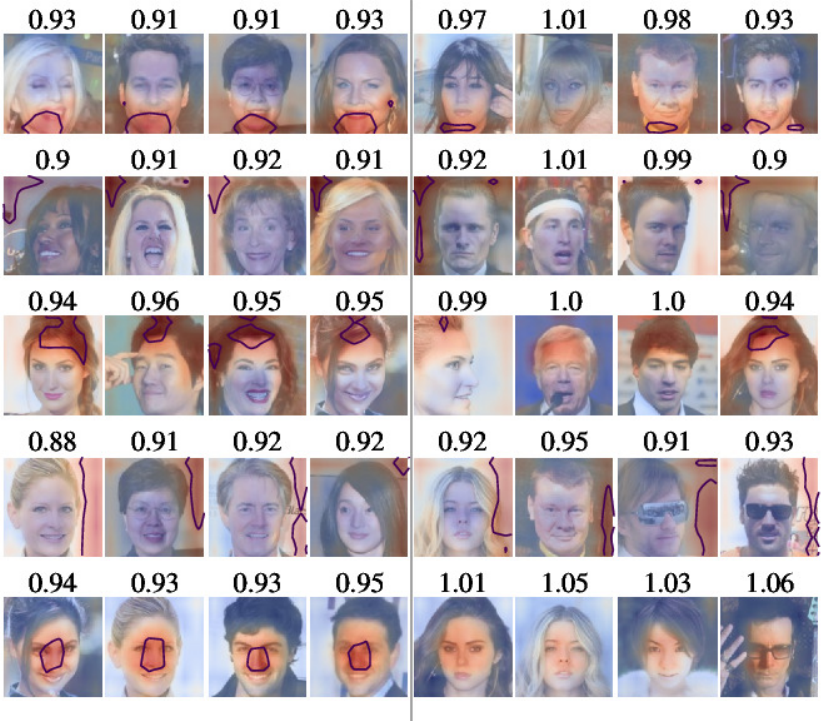}}	& &
		\multicolumn{2}{c}{\includegraphics[width=0.45\columnwidth]{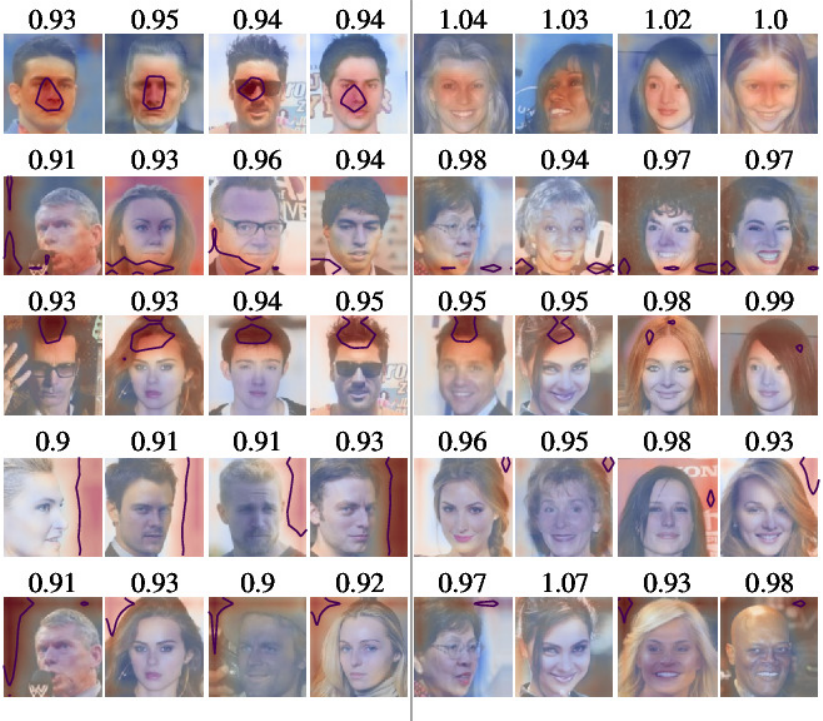}}	\\
	\end{tabular}
	
	\caption{SSCCD concepts on CelebA for a ResNet50 model for samples with the ``smiling'' (left) and ``not smiling'' (right) attributes. Concepts are ordered according to average concept relevance showing samples with smallest \textit{sample-concept proximity}. We apply the same smiling concept on a random selection of not smiling samples and vice-versa. Subtitles indicate the \textit{normalized sample-concept proximity} $\delta^{iA}/\delta^{A}$, i.e.\ samples with values $<1$ are (highly) activated. The ``smiling'' concepts can be roughly identified with chin/mouth, background, hair(line), background, nose and nose, background, hair, background, background in the case of ``not smiling''.}	
	\label{fig:celeba_smiling}
\end{figure}

To support the findings in \Cref{sec:celeba}, we show concepts for more attributes of the CelebA dataset in \Cref{fig:celeba_male} to \Cref{fig:celeba_smiling}. Additionally, we show concepts for the respective inverse attribute. These support the findings discussed in the main text, which we reiterate here for completeness: The identified concepts are highly specific both in terms of localization (c.f. chin for ``male'' vs.\ neck/hair for ``female'') and in terms of underlying features. The latter can be inferred from the fact that similarly localized concepts (e.g. nose for ``male''/``female'', hair for ``young/old'') are found for class and anti-class but evaluating a particular concept only weakly activates the corresponding region with the anti-concept.


\section{More concepts transfer experiments (Section 4.4)}
\begin{figure}[ht]
		\centering
	\def\arraystretch{0.5} 		
	\setlength\tabcolsep{5pt}		
	\begin{tabular}{cc}
	 street concept		& 		window concept 	\\
 	\includegraphics[width=0.48\columnwidth]{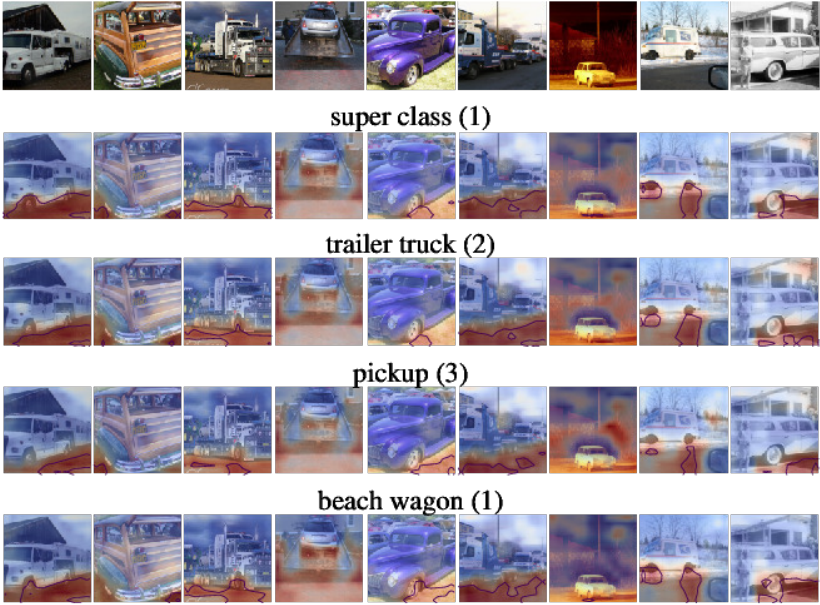}	&
		\includegraphics[width=0.49\columnwidth]{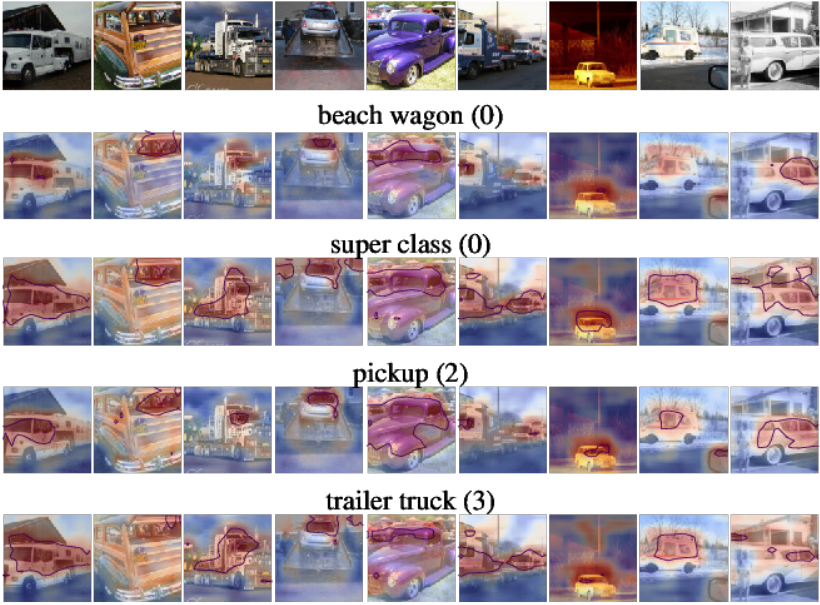} \\
	\end{tabular}
	\caption{Transferring and comparing the street/window concept from different vehicles classes on random test samples. Four most similar concepts according 
		\Cref{fig:concept_similarity}.	\label{fig:transfer_street_concept}
	}
\end{figure}

To further support the findings in \Cref{fig:concept_similarity}, we additionally show the transferability for two more concepts in \Cref{fig:transfer_street_concept}. Again, we find that all similar concepts are visually closely related and transfer in a sensible manner on all test samples.

\section{ACE concepts visualized on training set samples (Section 4.5)}
\begin{figure}[ht]
		\centering
	\includegraphics[width=1\columnwidth]{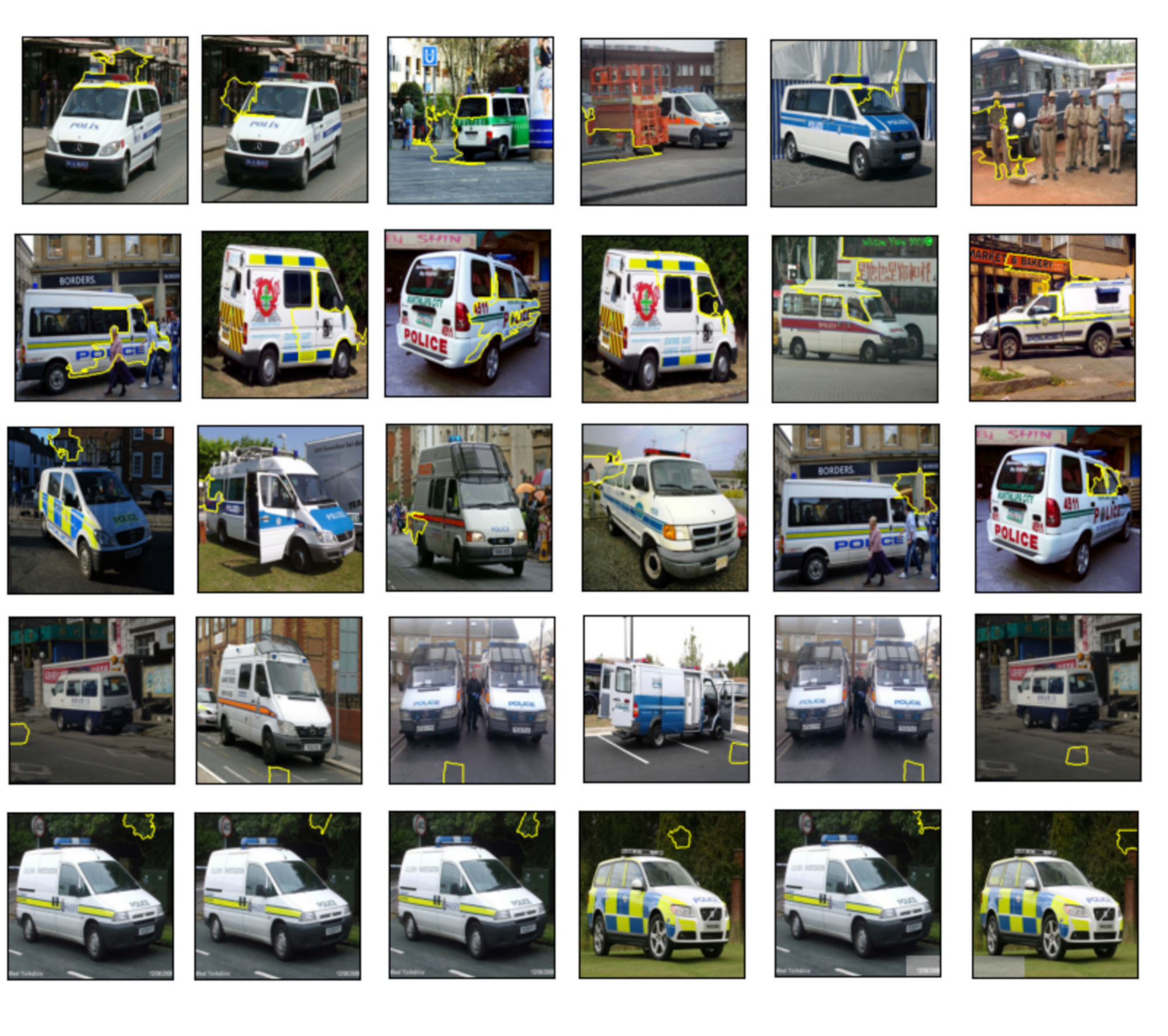}
	\caption{ACE concepts as shown in \Cref{fig:sdc_benchmark} visualized on training set samples. Each row corresponds to one concept. Like in \Cref{fig:sdc_benchmark}, we show the five most important concepts ordered by their TCAV scores from top row to bottom row. We can roughly identify background structures, police van body, windows, street and greenery.}	
	\label{fig:acetrain}
\end{figure}

Due to the difficulty of identifying coherent, meaningful structures in the ACE concepts visualized on validation set samples in \Cref{fig:sdc_benchmark} we show the same concepts on training set samples in \Cref{fig:acetrain}. Here, we can roughly identify background structures, police van body, windows, street and greenery. We suspect that generalization of concepts identified on training set samples to the test set is difficult for ACE.

\section{Rank-consistency of attribution methods for concept relevance}
The concept importance scores presented in this work are based on \ig{} attributions. We obtain per-sample concept relevances by summing over these attributions in the thresholded regions of the concept maps in input space. For class-wise concept relevance scores, we average over all per-sample scores in the class.
Here, we assess the effect of this relevance quantification method. Firstly, we replaced \ig{} with \pd{} and secondly, measure attributions in feature space directly (for both methods). For \ig{}, we use 200 steps and the dataset mean as a baseline. For \pd{}, we occlude each concept region using a train set imputer and 100 imputations. In \Cref{tab:attribution_consistency}, we show the median Spearman rank correlation between class-wise concept importance scores across 42 random classes from the \imagenet{} dataset. The concepts stem from the last convolutional layer of VGG16. While there are differences in the numerical values, the high median correlation coefficients in \Cref{tab:attribution_consistency} demonstrates that the concept ranking is consistent across all methods.

\begin{table}
	\centering
\setlength\tabcolsep{10pt}
\begin{tabular}{lrrrr}
\toprule
 &  \iga\,FS &  \iga \,IS &  \pd\, FS &  \pd\, IS \\
\midrule
\iga\, FS &              1.0 &            1.0 &              0.9 &            0.9  \\
\iga\, IS   &              1.0 &            1.0 &              0.9 &            0.9 \\
\pd \,FS &              0.9 &            0.9 &              1.0 &            0.9 \\
\pd\, IS   &              0.9 &            0.9 &              0.9 &            1.0 \\
\bottomrule
\end{tabular}
\caption{Median Spearman rank correlation coefficient between concept importance scores from \ig{} (\iga{}) and \pd{} in input space (IS) and feature space (FS) for concepts across 42 random classes from the \imagenet dataset and VGG16 (last convolutional layer).}
\label{tab:attribution_consistency}
\end{table}
\section{Source code and hyperparameter choices}
As part of this supplementary material, we provide source code to train custom models on CelebA data, to compute self-representations for a given dataset and pretrained model, to identify clusters via spectral clusters and to determine the corresponding bases for concept subspaces. Here, we provide basic usage information.
\subsection{Training a custom model on CelebA}
\begin{lstlisting}[basicstyle=\scriptsize, language=bash,caption={CelebA training call}]
#!/bin/bash
python imagenet_e2e.py --data path_to_celeba_data --dataset celeba
\end{lstlisting}
\subsection{Computing self-representation, clustering and concept subspaces}
In all our experiments, we fix the noise-controlling hyperparameter to $\gamma=10$. We choose $\tau=1.0$ for the hyperparameter corresponding to the elastic net penalty term. Lastly, we use two batches of 64 samples each and subsample features at a rate of $0.8$. Only for the final ResNet50 block, we use eight batches to account for the spatially smaller feature maps. Below, we provide an example for computing the self-representation, spectral clustering and subspace bases for the last convolutional layer (penultimate block) of VGG16 (ResNet50) and the ImageNet police van class.
\newpage
\begin{lstlisting}[basicstyle=\scriptsize, breaklines=True, language=python,caption={SSCCD usage}]
import concept_ssc
	
# SSCCD for the imagenet police van class 
# and the last convolutional layer (penultimate block) of VGG16 (ResNet50)
# (pretrained model from torchvision)
model_choice = 'vgg16'  # 'resnet50'
featurelayer_choice = 30  # 7
cssc = concept_ssc.ConceptSSC(ssc_method='elasticnet', n_concepts=5, gamma=10, tau=1.0, n_batches=2, batch_size=64, feature_subsample_ratio=0.8, featurelayer=featurelayer_choice,  model_arch=model_choice, dataset='imagenet', datapath='./', selected_classes=["police van",])
	 				  
# compute self-representation and spectral clustering 
cssc.fit(out_dir='./')
	
# determine bases for concept subspaces
bases = cssc.conceptspace_bases()
\end{lstlisting}

\end{document}